\documentclass[10pt,twocolumn,letterpaper]{article}

\usepackage[pagenumbers]{cvpr} %

\definecolor{cvprblue}{rgb}{0.21,0.49,0.74}
\usepackage[pagebackref,breaklinks,colorlinks,allcolors=cvprblue]{hyperref}

\usepackage{adjustbox}
\usepackage{graphicx}
\usepackage{caption}
\usepackage{subcaption}
\usepackage{printlen}
\usepackage{float}
\usepackage{wrapfig}
\usepackage{booktabs}
\usepackage{pifont}
\usepackage{makecell}
\usepackage{nicefrac}
\usepackage{url}
\usepackage{stmaryrd}
\usepackage{verbatim}
\usepackage{amsmath}
\usepackage{algorithm}
\usepackage{algpseudocode}
\usepackage{amsfonts}
\usepackage{pgf}
\usepackage{pgfplots}
\usepackage{soul}
\usepackage{tikz}
\usepackage{microtype}
\usepackage{framed}
\usepackage[capitalize]{cleveref}
\usetikzlibrary{angles,quotes,3d,math,arrows.meta,calc,positioning,fit,backgrounds,decorations.pathreplacing,calligraphy,shapes,shapes.multipart}

\newcommand{\cond}{\mathbf{c}}
\newcommand{\x}{\mathbf{x}}
\newcommand{\subjectj}{S_j}
\newcommand{\subject}[1]{S_{#1}}
\newcommand{\subjects}{\mathcal{S}}

\newcommand{\attributei}{A_i}
\newcommand{\attributes}{\mathcal{A}}
\newcommand{\attributescalei}{\lambda_i}
\newcommand{\expr}{\mathrm{expr}}
\newcommand{\xzero}{\mathbf{x}_0}

\newcommand{\eps}{\boldsymbol{\epsilon}}
\newcommand{\epspred}{\tilde{\eps}}
\newcommand{\epspredtheta}{\hat{\eps}_{\theta}}
\newcommand{\eye}{\mathbf{I}}
\newcommand{\loss}{\mathcal{L}}
\newcommand{\prompt}{P}
\newcommand{\promptp}{\prompt_+}

\newcommand{\emb}{\mathbf{e}}

\newcommand{\pembp}{\mathcal{E}_\mathrm{CLIP}(\promptp)}

\newcommand{\pembdiff}{\Delta\mathbf{e}}
\newcommand{\pembmod}{\mathbf{e}'}

\newcommand{\clip}{\mathrm{CLIP}}

\newcommand{\dclip}{d_\clip}

\newcommand{\cmark}{\ding{51}}
\newcommand{\xmark}{\ding{55}}

\newlength{\actuallinewidth}
\definecolor{ourgreen}{RGB}{46, 204, 113}
\definecolor{ourgreenborder}{RGB}{39, 174, 96}
\definecolor{ourblue}{RGB}{52, 152, 219}
\definecolor{ourblueborder}{RGB}{41, 128, 185}
\definecolor{ourorange}{RGB}{230, 126, 34}
\definecolor{ourorangeborder}{RGB}{211, 84, 0}
\definecolor{ourred}{RGB}{231, 76, 60}
\definecolor{ourredborder}{RGB}{192, 57, 43}
\definecolor{ouryellow}{RGB}{241, 196, 15}
\definecolor{ouryellowborder}{RGB}{243, 156, 18}
\definecolor{ourpurple}{RGB}{155, 89, 182}
\definecolor{ourpurpleborder}{RGB}{142, 68, 173}
\definecolor{ourturquoise}{RGB}{26, 188, 156}
\definecolor{ourturquoiseborder}{RGB}{22, 160, 133}
\definecolor{ourturquoise}{RGB}{26, 188, 156}
\definecolor{ourturquoiseborder}{RGB}{22, 160, 133}
\definecolor{ourwhite}{RGB}{236, 240, 241}
\definecolor{ourwhiteborder}{RGB}{189, 195, 199}
\definecolor{ourgray}{RGB}{149, 165, 166}
\definecolor{ourgrayborder}{RGB}{127, 140, 141}

\definecolor{ourwhite2}{RGB}{246, 247, 248}

\definecolor{ourhighlightcolor}{RGB}{46, 204, 113}
\newcommand{\ourhighlightcolorstr}{{\color{ourhighlightcolor}\textbf{green}}}

\newcommand{\shorttabular}[1]{\begin{tabular}{c}#1\end{tabular}}
\newcommand{\shorttabulara}[1]{\begin{tabular}{c}#1\\[-1mm]\\\end{tabular}}

\newcolumntype{H}{>{\setbox0=\hbox\bgroup}c<{\egroup}@{}}

\newcommand{\tikzstylenodedistance}{4mm}
\newcommand{\tikzstyleinnersep}{2mm}
\newcommand{\tikzstyleminimumheight}{8.75mm}
\newcommand{\tikzstyleminimumwidth}{12mm}

\tikzset{
    node distance=\tikzstylenodedistance,
    text centered,
    anchor=center,
}
\tikzset{
    standard node/.style n args={1}{%
        rectangle,
        rounded corners=0.1cm,
        fill=our#1,
        draw=our#1border,
        line width=0.04cm,
        minimum height=\tikzstyleminimumheight,
        minimum width=\tikzstyleminimumwidth,
        inner sep=\tikzstyleinnersep,
        text centered,
        anchor=center,
        align=center,
    }
}
\tikzset{
    standard node module/.style n args={0}{%
        rectangle,
        rounded corners=0.1cm,
        fill=ourturquoise,
        draw=ourturquoiseborder,
        line width=0.04cm,
        minimum height=\tikzstyleminimumheight, %
        minimum width=12mm, %
        inner xsep=\tikzstyleinnersep,
        inner ysep=1mm,
        text centered,
        anchor=center,
        align=center,
    }
}
\tikzset{
    standard node image/.style n args={1}{%
        rectangle,
        fill=our#1,
        draw=our#1border,
        line width=0.04cm,
        minimum height=\tikzstyleminimumheight,
        minimum width=\tikzstyleminimumwidth,
        inner sep=0,
        text centered,
        anchor=center,
        align=center,
    }
}
\tikzset{
    standard node circle/.style n args={1}{%
        fill=our#1,
        draw=our#1border,
        circle,
        inner sep=0.1cm,
        minimum height=0,
        minimum width=0,
    }
}
\tikzset{
    standard node circle/.prefix style = standard node
}

\tikzset{
    standard line/.style n args={0}{%
        line width=0.04cm,
        rounded corners=0.1cm,
    }
}
\tikzset{
    standard arrow/.style n args={0}{%
        -latex,
    }
}
\tikzset{
    standard arrow/.prefix style = standard line
}
\tikzset{
    standard double arrow/.style n args={0}{%
        latex-latex,
    }
}
\tikzset{
    standard double arrow/.prefix style = standard line
}

\tikzset{
    simple node image/.style n args={0}{%
        rectangle,
        inner sep=0,
        text centered,
        anchor=center,
        align=center,
        node distance=0mm
    }
}

\newcommand\rurl[1]{%
  \href{https://#1}{\nolinkurl{#1}}%
}

\def\paperID{2793} %
\def\confName{CVPR}
\def\confYear{2025}

\title{Continuous, Subject-Specific Attribute Control in T2I Models\\by Identifying Semantic Directions}

\author{
    Stefan Andreas Baumann$^{1}$, Felix Krause$^{1}$, Michael Neumayr$^{2}$, Nick Stracke$^{1}$, Melvin Sevi$^{3}$,\\
    Vincent Tao Hu$^{1}$, Björn Ommer$^{1}$\\
    $^{1}$CompVis @ LMU Munich, MCML, 
    $^{2}$TU Munich, 
    $^{3}$ENS Paris-Saclay\\
    {\tt\small \{stefan.baumann, b.ommer\}@lmu.de }\\
}

\begin{document}
\setlength{\linewidth}{\columnwidth}
\twocolumn[{
    \maketitle
    \vspace{-10mm} %

    \begin{center}
        \rurl{compvis.github.io/attribute-control}
    \end{center}
    \begin{center}
        \captionsetup{type=figure}
        \adjustbox{max width=\linewidth}{
            \hspace{-2mm}
\begin{tikzpicture}
    \node[] (center) at (0, 0) {};
    \node[above=of center,align=left] (prompt) {``A close-up photo of a \underline{man} and a \underline{woman} sitting on a bench''};
    \node[above=0mm of prompt,align=center] (a) {\textbf{(a) Our Method}: Detailed Control\\of \setul{0.5ex}{0.3ex}\setulcolor{ourwhiteborder}\ul{Localization} \textit{and} \setulcolor{ourblue}\ul{Expression}\setulcolor{black}};

    \node[standard node={white},below=of center,inner xsep=2mm,minimum width=0,fill opacity=.8,xshift=-7.13mm,yshift=16.25mm] (man) {\underline{man}\\[.5em]};
    \draw[standard line,latex-latex] ([xshift=1.5mm,yshift=-6.5mm]$(man.west |- man.north)$) -- node[midway] (slidermid) {$+$} node[pos=.75,yshift=-.1mm,scale=1.25] {\color{ourblue}$\bullet$} ([xshift=-1.5mm,yshift=-6.5mm]$(man.east |- man.north)$);
    \node[below=-1.5mm of slidermid,scale=.66] (deltaage) {$\Delta$age};
    
    \node[standard node={white},below=of center,inner xsep=.25mm,fill opacity=.8,xshift=10.3mm,yshift=16.25mm] (woman) {\underline{woman}\\[.5em]};
    \draw[standard line,latex-latex] ([xshift=1.5mm,yshift=-6.5mm]$(woman.west |- woman.north)$) -- node[midway] (slidermid) {$+$} node[pos=.42,yshift=-.1mm,scale=1.25] {\color{ourblue}$\bullet$} ([xshift=-1.5mm,yshift=-6.5mm]$(woman.east |- woman.north)$);
    \node[below=-1.5mm of slidermid,scale=.66] (deltaage) {$\Delta$age};

    \node[simple node image,below=7mm of prompt] (img) {\includegraphics[width=.67\linewidth]{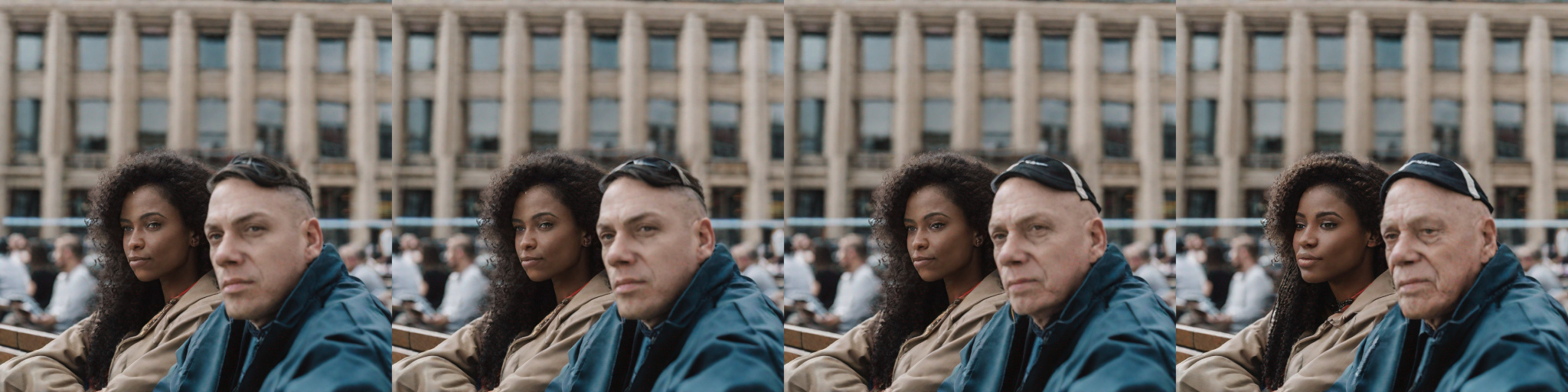}};
    \node[draw=ourhighlightcolor,line width=0.06cm,above=of img.south,minimum width=.1675\linewidth,minimum height=.1675\linewidth,xshift=-0.25125\linewidth,yshift=-4.05mm] {};
    \node[below=0mm of img,align=center,xshift=-44mm,scale=.66] (unmodifiedlabel) {Original Generated\\without Modifications};
    \draw[standard arrow] ([yshift=1.5mm]$(unmodifiedlabel.east |- unmodifiedlabel.east)$) -- node[below,scale=.66] {Gradually Stronger Age Modulation Applied} ([yshift=1.5mm]$(img.east |- unmodifiedlabel.east)$);

    \node[right=39mm of a,align=center,yshift=8.5mm] (b) {\textbf{(b) Concept Sliders} \citep{gandikota2023sliders}: Only \textit{Global} Control};
    \node[simple node image,below=3.5mm of b.west,xshift=.337\linewidth] (imgconcept) {\includegraphics[width=.67\linewidth]{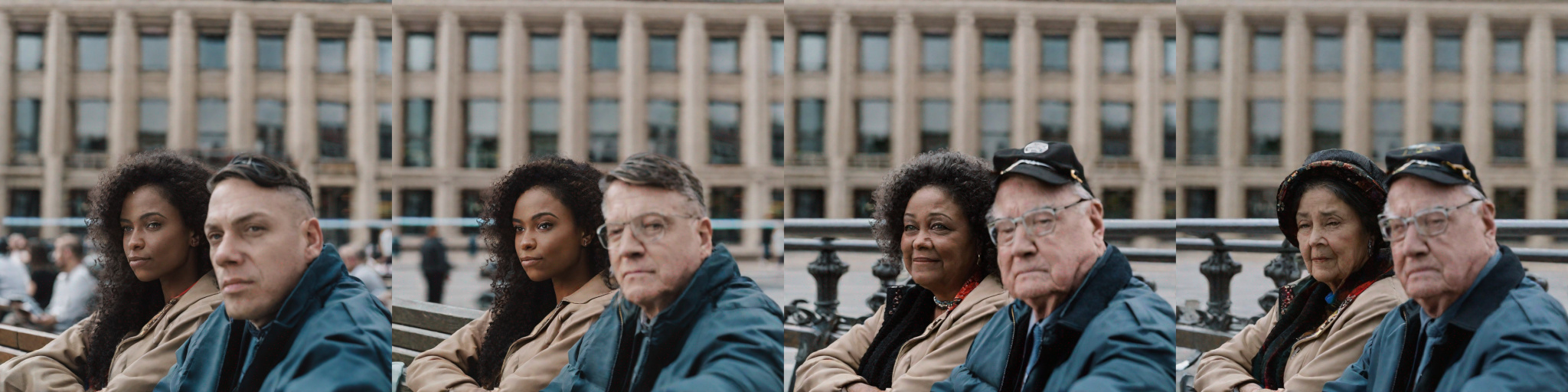}};
    \node[draw=ourhighlightcolor,line width=0.06cm,above=of imgconcept.south,minimum width=.1675\linewidth,minimum height=.1675\linewidth,xshift=-0.25125\linewidth,yshift=-4.05mm] {};
    \node[below=0mm of imgconcept,align=center,xshift=-44mm,scale=.66, draw=none] (unmodifiedlabelconcept) {Original Generated\\without Modifications};
    \draw[standard arrow] ([yshift=1.5mm]$(unmodifiedlabelconcept.east |- unmodifiedlabelconcept.east)$) -- node[below,scale=.66] {Gradual \underline{Global} Age Increase} ([yshift=1.5mm]$(imgconcept.east |- unmodifiedlabelconcept.east)$);
    
    \node[simple node image,below=13mm of imgconcept] (imgp2p) {\includegraphics[width=.67\linewidth]{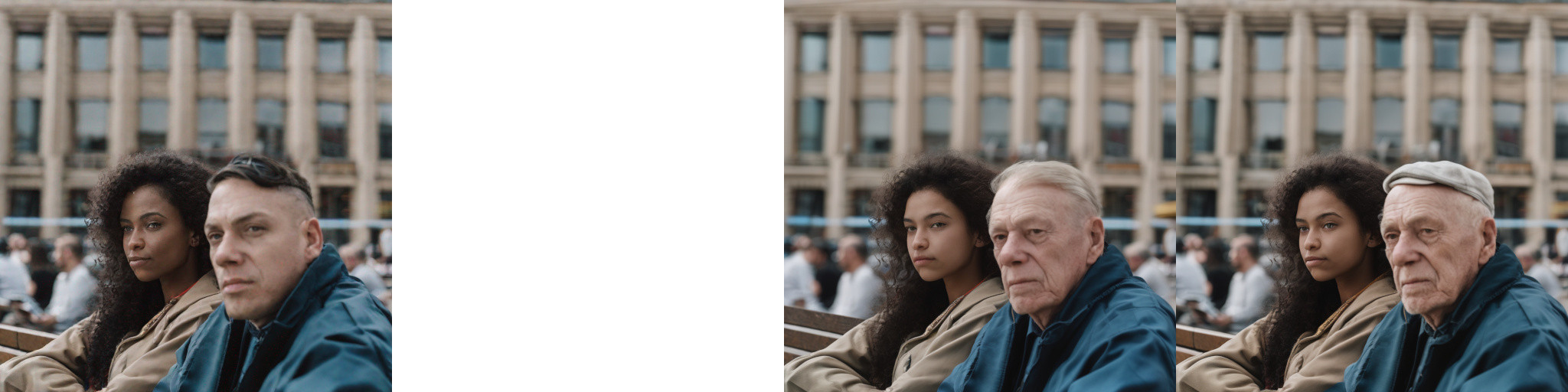}};
    \node[above=-2mm of imgp2p.south,xshift=-15mm,fill=white,inner sep=.5mm,scale=.66] {\color{ourred}$\boldsymbol{\times}$ continuity breaks};
    \node[draw=ourhighlightcolor,line width=0.06cm,above=of imgp2p.south,minimum width=.1675\linewidth,minimum height=.1675\linewidth,xshift=-0.25125\linewidth,yshift=-4.05mm] {};
    \node[above=.7mm of imgp2p,align=center,xshift=-20mm] (c) {\textbf{(c) Prompt-to-Prompt} \citep{hertz2023prompttoprompt}: No \textit{Fully Continuous} Control};
    \node[below=0mm of imgp2p,align=center,xshift=-44mm,scale=.66,] (unmodifiedlabelp2p) {Original Generated\\without Modifications};
    \node[below=0mm of imgp2p,align=center,xshift=44mm,scale=.66,] (unmodifiedlabelp2p2) {Image Generated\\with Modified Prompt};
    \draw[standard arrow] ([yshift=0.33mm]$(unmodifiedlabelp2p.east |- unmodifiedlabelp2p.east)$) -- node[,fill=white,rotate=0,inner sep=-2.5pt,outer sep=0]{\color{ourred}$\boldsymbol{\times}$} node[below,scale=.66] {No Gradual, Fine-Grained Control} ([yshift=0.33mm]$(unmodifiedlabelp2p2.west |- unmodifiedlabelp2p2.west)$);
\end{tikzpicture}
\hspace{-4mm}

        }
        \captionof{figure}{\textbf{(a)} We augment the prompt input of image generation models with \textit{fine-grained control} of attribute expression in generated images (unmodified images are marked in \ourhighlightcolorstr) in a \textit{subject-specific} manner \textit{without additional cost} during generation. \textbf{(b, c)} Previous methods only allow \textit{either} fine-grained expression control or fine-grained localization when starting from the image generated from a basic prompt.}
        \label{fig:teaser}
    \end{center}
}]
\setlength{\linewidth}{\columnwidth}

\begin{abstract}

Recent advances in text-to-image (T2I) diffusion models have significantly improved the quality of generated images. However, providing efficient control over individual subjects, particularly the attributes characterizing them, remains a key challenge. While existing methods have introduced mechanisms to modulate attribute expression, they 
typically provide either detailed, object-specific localization of such a modification or full-scale fine-grained, nuanced control of attributes. 
No current approach offers both simultaneously, resulting in a gap when trying to achieve precise continuous and subject-specific attribute modulation in image generation.
In this work, we demonstrate that token-level directions exist within commonly used CLIP text embeddings that enable fine-grained, subject-specific control of high-level attributes in T2I models. We introduce two methods to identify these directions: a simple, optimization-free technique and a learning-based approach that 
utilizes the T2I model
to characterize semantic concepts 
more specifically. Our methods allow the augmentation of the prompt text input, enabling fine-grained control over multiple attributes of individual subjects simultaneously, without requiring any modifications to the diffusion model itself. This approach offers a unified solution that fills the gap between global and localized control, providing competitive flexibility and precision in text-guided image generation.

\end{abstract}

\section{Introduction}
Text-to-image (T2I) diffusion models have rapidly advanced, achieving remarkable quality in generating visually stunning images \citep{rombach2022high,imagenteamgoogle2024imagen3}. However, as the quality of generated images improves, the need for precise control over the generation process becomes increasingly crucial. This control should extend beyond simply adjusting \emph{what} is depicted in the scene. It must also provide nuanced control of the attributes describing \emph{how} these objects are characterized. Attributes, such as a person’s age, are not binary or static -- they often span a continuum, requiring models to capture fine-grained variations to produce results that align with user intent.

Currently, a fundamental gap exists: no method provides fine-grained modulation and subject-specific localization simultaneously. Recent works like Prompt-to-Prompt (P2P)~\citep{hertz2023prompttoprompt} and Concept Sliders~\citep{gandikota2023sliders} have made significant strides in introducing control into T2I models. P2P enables localized expression changes, allowing adjustments to specific aspects of a given image based on text modifications but only allow a very limited range of modulation. Concept Sliders facilitate fine-grained modulation over global attributes across all subject instances. This limitation means that while we can tweak attributes globally or localize changes to subjects, we still lack a unified, generalized approach capable of concurrently achieving fine-grained control for both aspects.

This work aims to bridge this gap by introducing a method that enables unified, subject-specific, fine-grained control over attributes within T2I diffusion models. Unlike existing methods that provide either localized coarse control or global fine-grained control, our approach offers precise modulation of attributes that can be directed at specific subjects within the generated image (see \cref{fig:teaser}). This results in an unprecedented level of intuitive control, allowing users to finely tune not just what appears in an image but how it appears, down to the smallest level of attribute expression.

\noindent We summarize our main contributions as follows:
\begin{itemize}
    \item We show that token-level edit directions exist within common CLIP embeddings, enabling fine-grained control of subject-specific attributes, and show that diffusion models can effectively interpret these directions.
    \item We introduce a simple, optimization-free approach to identify attribute-specific directions by contrasting text prompts that describe the desired attributes or concepts.
    \item We introduce a second, learning-based method that identifies more robust directions through backpropagation of high-level semantic concepts to the text embedding input, using a reconstruction loss objective.
    \item We show that these token-level edit directions enable fine-grained, subject-specific, compositional control of attributes and concepts in generated images.
\end{itemize}

\section{Related Work}

The rapid advancements in generative models for image and video synthesis, particularly diffusion models like Stable Diffusion~\citep{rombach2022high}, have spurred efforts to develop techniques for fine-grained editing and control of specific attributes in generated content. Our work focuses on enabling precise, subject-specific control in images by targeting individual characteristics in a controlled and continuous manner.

Existing methods for controlled generation and image editing can be broadly categorized based on the underlying generative models -- primarily Generative Adversarial Networks (GANs)~\citep{goodfellow2014generative} and Diffusion Models~\citep{ho2020denoising} --, and the mechanisms they use for control -- typically latent space manipulations or textual descriptions.

\paragraph{T2I Diffusion Model Preliminaries}
T2I Diffusion models~\citep{rombach2022high,podell2024sdxl} simulate a reverse diffusion process $p_\theta(\x_{0:T}|\prompt)$ that enables sampling from the distribution of images $p_\theta(\x_{0}|\prompt)$ given a text conditioning $\prompt$ and a Gaussian noise sample $\x_T$. They iteratively denoise $\x_T$ using a diffusion model $\epspredtheta(\x_t|\prompt,t)$. This is typically done by learning to predict the noise content $\eps$ in the sample $\x_t = \alpha_t \xzero + \sigma_t \eps$ using the following loss function:
\begin{multline}\label{eq:diffusion_reconstruction}
    \loss_\mathrm{Diffusion} = \mathbb{E}_{(\x_0, \cond) \sim p_\text{data}(\x_0, \cond), \eps \sim \mathcal{N}(0,\eye),t \sim \mathcal{U}(0, T]} \\
    w(t) \left\| \eps - \epspredtheta(\alpha_t \xzero + \sigma_t \eps \,|\, \cond, t) \right\|_2^2,
\end{multline}
where $\epspredtheta(\cdot)$ is the diffusion model conditioned on the timestep $t$ and the conditioning signal $\cond$, $w(t)$ is a loss weighting term, and $\alpha_t$ and $\sigma_t$ are noise schedule parameters. The conditioning $\cond$ is typically obtained using a CLIP~\citep{clip} text encoder $\mathcal{E}_\mathrm{CLIP}$ as a tokenwise embedding $\emb = \mathcal{E}_\mathrm{CLIP}(\prompt)$ of a text prompt $\prompt$.

\subsection{GAN-based Image Editing and CLIP-Based\\Directions} 

GANs \citep{goodfellow2014generative, radford2015dcgan}, particularly StyleGANs \citep{karras2019style}, are popular for image editing due to their generative power and disentangled latent space. Methods like InterFaceGAN \citep{shen2020interfacegan} manipulate attributes by identifying latent space directions. Approaches such as StyleCLIP \citep{patashnik2021styleclip}, CLIP2StyleGAN \citep{abdal2022clip2stylegan}, and TediGAN~\citep{xia2021tedigan} use CLIP \citep{clip} for text-based guidance in latent space editing.
Despite these advancements, these methods inherit the limitations of StyleGAN and struggle to generalize to complex, multi-subject images.

\subsection{Steering the Diffusion Generation Process}\label{sec:related_diffusion_control}

\paragraph{Direction-based Control}

Similar to GAN-based editing, approaches like DiffusionCLIP \citep{diffusionclip} use CLIP for editing with unconditional closed-domain diffusion models. Recent methods, such as Asyrp~\citep{kwon2023diffusion}, InterpretDiffusion~\citep{li2024self}, LFM~\citep{hu2024latentspaceediting}, and BoundaryDiffusion~\citep{zhu2023boundaryguided}, modulate learned directions in the diffusion backbone or noise space, similar to StyleGAN. Concept Sliders~\citep{gandikota2023sliders} achieve disentangled attribute modulation by training attribute-specific LoRAs~\citep{hu2021lora}, however, these methods typically lack subject-specificity, as they perform global modulations.
Mask-based approaches like MAG \citep{mao2023mag} allow more targeted control but require significant user input to define the masks. 

\begin{figure}[t]
    \centering
    \begin{minipage}{\linewidth}\hspace{-3mm}
        \vspace{-5mm}
        \input{figs/continuousness_interpolation/figure}
    \end{minipage}
    \vspace{-2mm}
    \caption{The tokenwise CLIP text embedding space is not globally smooth. We linearly interpolate between the embeddings of two prompts while keeping the noise seed fixed. Near the original embeddings, changes are smooth and semantically interpretable, but strong phase transitions exist between substantially different subjects (e.g., ``car'' vs.\ ``frog'').}
    \label{fig:continuousness_interpolation}
\end{figure}

\paragraph{Attention Map-based Control}
Building on the observation by \cite{hertz2023prompttoprompt} that fixing attention maps during generation while changing the text prompt enables generating variations of images, a range of control methods utilizing this mechanism have been introduced. Methods like Prompt-to-Prompt~\citep{hertz2023prompttoprompt}, MasaCtrl~\citep{cao2023masactrl}, AdapEdit~\citep{ma2024adapedit}, and many others~\citep{brooks2023instructpix2pix,simsar2023lime,zhang2023hive,li2024zone,mirzaei2024watch} leverage attention control combined with prompt editing to allow subject-specific manipulations via text. These methods provide intuitive control and subject-specificity but suffer from the inherent discreteness of text inputs and struggle with fine-grained control over the magnitude of changes, offering partial magnitude control at best.

\paragraph{From Controlled Generation to Editing}
Inversion techniques are employed to map images back into a model's latent space for editing real images. In GAN-based methods, Image2StyleGAN \citep{abdal2019image2styleganembedimagesstylegan} and In-Domain GAN Inversion \citep{zhu2020indomainganinversionreal} are commonly used. Similarly, for diffusion models, methods like DDIM Inversion \citep{dhariwal2021adm}, Null-Text Inversion \citep{mokady2023nulltextinversion}, ReNoise \citep{garibi2024renoise}, and others \cite{huberman2024edit,deutch2024turboedit,wu2024turboedit,tsaban2023ledits} map images to the latent noise space and enable editing via re-generation with changed conditioning or guidance. As our method augments the diffusion model's text input with additional control, it can be combined with inversion methods to perform real image editing.

\section{Method}

Let $M$ denote the number of attributes we consider in our work and let $N$ denote the number of subjects mentioned in a prompt $\prompt$, $\attributes = \left\{ \attributei \mid i \in \llbracket 1, M \rrbracket \right\}$ denote the set of attributes $\attributei$ and $\subjects_\prompt = \left\{ \subjectj \mid j\in \llbracket 1, N \rrbracket \right\}$ denote the set of subjects $\subjectj$ mentioned in the prompt.
We aim to influence the generation process to enable control over the expression $\expr(\attributei)$ (i.e., how strongly it is present) of specific attributes $\attributei \in \attributes$ of specific subjects $\subjectj \in \subjects_\prompt$. As an example, consider the prompt ``a portrait of a man and woman sharing a laugh''. If the man should be younger, one can change ``man'' to ``young man'', but this does not offer continuous control over how young the man is supposed to be. Instead, we aim to provide the same subject-specificity that changing the prompt offers, but without the limitations of the non-continuousness of language.
Unlike previous works, we wish to provide control that is simultaneously i) continuous, ii) subject-specific, and iii) does not require manual image masks or reference images.

Our key observation is that the diffusion model's \textit{interpretation} of the tokenwise CLIP text embedding vector $\emb = \mathcal{E}_\mathrm{CLIP}(\prompt) = (\emb_\mathrm{<SOS>}, \emb_1, \dots, \emb_k, \emb_\mathrm{<EOS>})$, which is typically used to condition the model, is \textit{locally} smooth and enables \textit{subject-specific} semantic modulations (\cref{sec:tokenwise_clip_space_diffusion_interpretation}). Using this property, we can continuously modulate semantic attributes of specific subject instances in the prompt $\prompt$. To enable targeted modulation of specific attributes, we introduce methods to identify latent space directions corresponding to attributes $\attributes$ (e.g., ``old", ``happy", ``expensive").

\begin{figure}[t]
    \centering
    \begin{minipage}{.7\columnwidth}
        \input{figs/random_subject_specific_deviation/figure}\vspace{-6mm}
    \end{minipage}
    \caption{The tokenwise CLIP embedding space enables subject-specific interventions. Changes to the embedding of subject tokens can lead to disentangled local changes focused on that subject.}
    \label{fig:random_subject_specific_deviation}
\end{figure}

\subsection{Interpretation of tokenwise CLIP\\Text Embeddings in Diffusion Models}\label{sec:tokenwise_clip_space_diffusion_interpretation}
\paragraph{Global \textit{v.s.}\ Local Behavior}
Unlike the \textit{pooled} text embedding space of CLIP \citep{clip} models, which has been explored extensively in previous works \citep{patashnik2021styleclip,wang2022learningdecomposevisualfeatures,ramesh2022hierarchicaltextconditionalimagegeneration,zhuang2024magnet,wu2024relation} and found to contain global image information, the \textit{tokenwise} text embedding space has not been investigated as much. Previous methods \citep{chefer2024hiddenlanguage,li2024getwhatyouwant,wang2024concept} typically also interpret this space \textit{globally}, applying projections onto subspaces to decompose concepts or eliminate them from the generated images.
Conversely, we find two distinct \textit{local} behaviors in the tokenwise CLIP embedding space as interpreted by diffusion models~\citep{podell2024sdxl}. We can observe strong local (w.r.t. embedding space) phase changes when interpolating between substantially different subjects (see \cref{fig:continuousness_interpolation}, top). Here, minor changes cause drastic image changes. At the same time, the space shows smooth, semantically interpretable changes in the vicinity of the original embeddings and when interpolating between similar subjects (see \cref{fig:continuousness_interpolation}, bottom).

\paragraph{Subject-Specificity}
The CLIP tokenizer typically maps individual words to single tokens. Diffusion models also directly attend to adjectives added to subjects in the prompt to determine details of the subjects' appearance \citep{hertz2023prompttoprompt,rassin2023linguistic}. Despite this direct connection, additional information is also stored in other tokens, especially other tokens describing the subject, and is interpreted by the diffusion model~\citep{li2024getwhatyouwant}. Our key observation here is that we can exploit this semantic aggregation in the subject tokens to perform targeted interventions: modulating the token embedding $\emb_{[\subjectj]}$ of a specific subject $\subjectj$ primarily affects only that subject in the generated image (see \cref{fig:random_subject_specific_deviation}), without the need for adding new tokens.

\subsection{Identifying Semantic Directions from\\Contrastive Prompts}\label{sec:method_naive_deltas}
To use the key observations in \cref{sec:tokenwise_clip_space_diffusion_interpretation} for subject-specific control, we have to identify which directions enable modulating specific attributes. We previously found that interpolation of the tokenwise text embeddings leads to locally smooth changes around the original embeddings (c.f. \cref{fig:continuousness_interpolation}).
Motivated by this finding, we propose identifying semantic directions in the tokenwise embedding space by comparing embeddings of contrastive prompts.

Formally, given a target attribute $\attributei$, defined via an adjective (e.g., \textit{``old''}), we want to identify a direction vector $\pembdiff_{\attributei} \in \mathbb{R}^{\dclip}$ that can be added to the embedding of a target subject token $\emb_{[\subjectj]}$ to modulate the expression of that attribute $\expr_{\subjectj}(\attributei)$ in the generated image.
To identify this direction, we first obtain the tokenwise CLIP embeddings for two prompts: a neutral prompt $\prompt$ describing a single subject $\subject{}$ and a positive prompt $\prompt_+$, which prepends the adjective to the subject, resulting in a contrastive pair. Then, we compute the difference between the subject token embeddings $\emb_{[\subject{}]}$:
\begin{equation}
    \pembdiff_{\attributei} = (\mathcal{E}_\mathrm{CLIP}(\prompt_+) - \mathcal{E}_\mathrm{CLIP}(\prompt))_{[\subject{}]}.
\end{equation}
This directly yields a direction $\pembdiff_{\attributei}$ that captures the change induced by prepending the adjective to the subject noun in the text prompt. To obtain more robust estimates of this direction, we average it over a multitude of prompt pairs which describe the same target attribute $\attributei$.

To modulate that attribute's expression $\expr_{\subjectj}(\attributei)$ in the generated image for a given prompt embedding $\emb$ and target subject $\subjectj$, we apply the modulation $\attributescalei\pembdiff_{\attributei}$ to $\emb$ with
\begin{equation}\label{eq:attribute_modulation}
    \pembmod(\emb,\attributescalei\pembdiff_{\attributei})_{[\subjectj]} = \emb_{[\subjectj]} + \attributescalei\pembdiff_{\attributei},
\end{equation}
where $\attributescalei$ is a scalar controlling the magnitude of the modulation. This modified embedding is then passed to the diffusion model in place of $\emb$. This omits any changes to tokens other than the target subject noun, including the \texttt{<EOS>} token, which plays a crucial role in the image generation process~\citep{yesiltepe2024mist,li2024getwhatyouwant,wu2024relation}. Despite this, it successfully enables the modulation of target attributes (see \cref{fig:naive_vs_learned}a).

\begin{figure}[t]
    \centering
    \adjustbox{max width=\linewidth}{
        \input{figs/naive_vs_learned/figure}
    }
    \caption{
Variations along ``vehicle price'' directions identified using our methods. (a) Modulate along direction from difference-based approach (\cref{sec:method_naive_deltas}). (b) Modulate along direction from robust learned approach (\cref{sec:method_robust_deltas}). Unmodified images are marked in \ourhighlightcolorstr. These directions successfully capture the target attribute and allow for fine-grained modulation but (a) also shows unwanted side-effects such as flipping the car's orientation.}
    \label{fig:naive_vs_learned}
\end{figure}

\subsection{Identifying Robust Semantic Directions\\via Diffusion Noise Predictions}\label{sec:method_robust_deltas}

\begin{figure*}[t]
    \centering
    \input{figs/manifold_illustration/figure}
    \vspace{-1.5mm}
    \caption{Illustration of our method's intuition. We find that directions corresponding to modulating an attribute $\attributei$ in the noise prediction space $\Delta \epspred$ ({\color{ourgreen}\textbf{green}}) from a specific starting point $\x_t$ can be backpropagated ({\color{ourpurple}\textbf{purple}}) through the diffusion model (\cref{eq:loss_direction}) to obtain a generalized matching direction $\pembdiff_{\attributei}$ ({\color{ourblue}\textbf{blue}}) in the tokenwise embedding space. $\mathcal{E}(\prompt)$ is the prompt embedding, $\epspredtheta(\cdot)$ the diffusion model.}
    \label{fig:noise_space_token_emb_space_distillation_illustration}
\end{figure*}

Although the simple difference-based method introduced in \cref{sec:method_naive_deltas} is effective in many scenarios, it has several limitations. In practice, it often leads to unintended side effects (see \cref{fig:naive_vs_learned}) and is limited to attributes $\attributei$ expressible as prefixes to the subject noun, due to the causality of the CLIP text encoder. To address these issues, we propose a substantially more robust approach for identifying such directions.
To obtain more robust directions, we use a T2I diffusion model to identify associations of adjectives to directions in the tokenwise embedding space. This effectively inverts the typical relation, where language models are used to augment the T2I model, such as with prompt augmentation \citep{betker2023dalle3}. We use the diffusion model to identify sample-specific directions corresponding to modulations of the target attribute in the noise prediction space and backpropagate them through the diffusion model to discover \textit{generalizable}, \textit{fine-grained} local modulation directions $\pembdiff_{\attributei}$ within the tokenwise CLIP embedding space. Specifically, we aim to apply the modulation and change the image similarly to adding an adjective to the prompt, but without adding additional tokens or affecting the rest of the embedding, and while enabling fine-grained modulations.

We start with a random (generated) image $\xzero$ and its corresponding neutral prompt $\prompt$ describing one subject $\subject{}$ and sample a random timestep $t \sim \mathcal{U}[0, T)$. We obtain the noised latent as $\x_t = \alpha_t \xzero + \sigma_t \eps, \eps \sim \mathcal{N}(0, \mathbf{I})$,
where $\alpha_t$ and $\sigma_t$ are time-dependent noise schedule coefficients. Then, we predict the noise for two different prompts with the T2I diffusion model: the original prompt, $\epspred = \epspredtheta(\x_t|\prompt)$ and the prompt with the adjective added, $\epspred_+ = \epspredtheta(\x_t|\promptp)$. Using these two noise predictions, we obtain a direction $\Delta\epspred = \epspred_+ - \epspred$ in that particular image's and prompt's noise space corresponding to modulating $\attributei$.\footnote{If an attribute can be described using a antonym pair of adjectives (e.g., \textit{``old''} and \textit{``young''}), we use the direction $\Delta\epspred = \epspred_+ - \epspred_-$ instead.}
Finally, we distill that direction in the noise space through the diffusion model into the direction $\pembdiff_{\attributei}$ (see \cref{fig:noise_space_token_emb_space_distillation_illustration} for an illustration) using the reconstruction loss
\begin{multline}\label{eq:loss_direction}
    \loss(\xzero, \emb; \pembdiff_{\attributei}) = \mathbb{E}_{\attributescalei, \eps \sim \mathcal{N}(0,\eye),t \sim \mathcal{U}[0, T)} \\
    w(t) \left\| (\eps + \attributescalei\Delta\epspred) - \epspredtheta(\x_t \,|\, \pembmod(\emb, \attributescalei \pembdiff_{\attributei}), t) \right\|_2^2.
\end{multline}
adapted from \cref{eq:diffusion_reconstruction}. To capture the full scale of potential changes, including fine-grained ones, we randomly vary $\attributescalei$. Finally, to obtain a robust, generalizable direction for $\attributei$, we optimize $\pembdiff_{\attributei}$ using AdamW \citep{loshchilov2018decoupled} over a wide range of different sampled images $\xzero$ from different base prompts $\prompt$, noises $\eps$, and timesteps $t$. Unlike \cite{gandikota2023sliders}, we predict a continuous target direction and train on that by continuously varying $\attributescalei$. We provide an overview of the full training algorithm in \cref{alg:semantic_directions}.

\subsection{Attribute Control}
\begin{figure}
    \centering
    \captionsetup{format=plain}
    \adjustbox{max width=.3\textwidth}{\input{figs/car_price_clip_diffs.pgf}}
    \vspace{-2mm}
    \caption{Applying modulations $\attributescalei\pembdiff_{\attributei}$ gradually shifts the distribution of generated images w.r.t. the expression of the target attribute $\expr(\attributei)$. We show the kernel density estimation of the CLIP score difference between \textit{``a photo of an expensive car''} \& \textit{``a photo of a car''} (original prompt) while modulating $\expr_\mathrm{car}(\mathrm{vehicle\ price})$.}
    \label{fig:attribute_distribution_shift_kde}
\end{figure}
During inference time, we use \cref{eq:attribute_modulation} to control the expression $\expr_{\subjectj}(\attributei)$ of an attribute $\attributei$ of a specific subject $\subjectj$. By adding the modulation $\pembdiff_{\attributei}$ to the target subject $\subjectj$ in the tokenwise prompt embedding $\emb$, we bias the distribution of generated images $p(\xzero)$ towards increased or decreased expression of the target attribute $\attributei$ for the target subject $\subjectj$ (see \cref{fig:attribute_distribution_shift_kde}). We typically apply the modulation after the first 20\% of sampling steps to achieve more fine-grained changes, as in \citep{meng2021sdedit,gandikota2023sliders}. Moreover, this approach supports the additivity of attribute modulations, allowing for multiple simultaneous edits. By adding several modulation vectors $\pembdiff_{\attributei}$, we can independently adjust different attributes for the same subject $\subjectj$ without interfering with each other. Our method also allows for editing multiple subjects within the same image by applying separate modulations to different subjects. As applying our method only requires one addition, it effectively adds zero inference cost.

\paragraph{Application to Real Image Editing}
In addition to modulating attributes in generated images, our method can also be used to perform fine-grained edits of real images. We first invert the given real image $\mathcal{I}$ with a matching caption (obtained, e.g., by user input or synthetic captioning) into its corresponding noise latent $\x_T$ using an off-the-shelf inversion method \citep{garibi2024renoise}. Then, we regenerate the image while applying our attribute modulation to the target subject in the same manner as when generating images from scratch to obtain fine-grained subject-specific edits of real images.

\section{Experiments}\label{sec:experiments}

\begin{figure*}[t]
    \centering
    \adjustbox{max width=\linewidth}{
    \begin{minipage}{\linewidth}
        \begin{minipage}{.58\textwidth}
            \centering

            \adjustbox{max width=\linewidth}{
            \begin{tabular}{lHH@{}c@{}HHccc@{}r}
                & & & \multicolumn{1}{c}{\textbf{(a)} Subject-Specificity} & & & \multicolumn{2}{c}{\textbf{(b)} Disentangledness} & \textbf{(c)} & \textbf{(d)} Performance\\
                \midrule
                \textbf{Method} &
                 &  & \textbf{Subject-Specificity} $\uparrow$ &
                 &  &
                $\Delta\mathbf{Id}$ ${\downarrow}$ & $\mathbf{LPIPS}$ ${\downarrow}$ &
                \textbf{Continuous} & \textbf{Time} $\downarrow$\\
                \midrule
                Adjectives in Text Prompt & - & - & {{4.14}} & - & - & 0.48 & 0.28 & \xmark & \underline{12.0s} [4.17it/s]\\
                \rowcolor{ourwhite2}
                Concept Sliders \citep{gandikota2023sliders} & 0.997 & 1.014 & \xmark & {0.068} & 0.028 & 0.45 & 0.20 & \cmark & 33.8s [1.48it/s] \\ 
                Prompt-to-Prompt \citep{hertz2023prompttoprompt} & 0.974 & 1.118 & 3.93 & 0.091 & 0.039 & 0.60 & 0.29 & $\sim$\xmark\phantom{$\sim$} & 23.5s [4.16it/s] \\ 
                \rowcolor{ourwhite2}
                AdapEdit \citep{ma2024adapedit} & {0.183} & {0.068} & 
                \textbf{6.92} & {} & {} & \underline{0.24} & {0.10} & \xmark & 13.2s [7.58it/s] \\ %
                MasaCtrl (Gen.) \citep{cao2023masactrl} & 0.935 & \textbf{0.502} & 2.48 & 0.117 & 0.054 & 0.66 & 0.28 & {\xmark} & 153.0s [0.65it/s] \\
                \rowcolor{ourwhite2}
                MasaCtrl (Edit*) \citep{cao2023masactrl} & {0.964???} & 0.849 & 1.93 & 0.121 & 0.073 & 0.61 & 0.43 & {\xmark} & \textbf{10.2s} [4.86it/s]  \\
                Ours & \textbf{0.521} & {0.629} & 3.35 & 0.072 & {0.019} & 0.40 & 0.10 & \cmark & \underline{12.0s} [4.17it/s] \\
                \rowcolor{ourwhite2}
                Ours + Prompt-to-Prompt \citep{hertz2023prompttoprompt} & 0.947 & \underline{0.578} & 2.23 & \textbf{0.042} & \textbf{0.010} & {0.37} & \underline{0.08} & \cmark & 23.5s [4.16it/s]\\
                Ours + AdapEdit \citep{ma2024adapedit} & - & - & \underline{6.46} & - & - & \textbf{0.19} & \textbf{0.05} & \cmark & 13.2s [7.58it/s] \\
                \rowcolor{ourwhite2}
                Ours + ReNoise \citep{garibi2024renoise} & - & - & 2.28 & - & - & 0.82 & 0.32 & \cmark & 32.2s [5.367it/s]\\
                \midrule
                \textit{Ablations} (see \cref{sec:additional_ablations} for an extended version) \\
                \rowcolor{ourwhite2}
                Ours (w/o Delay) & {0.824} & 0.696 & 3.47 & 0.091 & 0.041 & 0.50 & 0.22 & \cmark & {12.0s} [4.17it/s] \\
                Our CLIP Difference Method (\cref{sec:method_naive_deltas}) & - & - & 2.38 & - & - & 1.20 & 0.58 & \cmark & {12.0s} [4.17it/s] \\
                \rowcolor{ourwhite2}
                Directly modulating $\Delta\epspred$ (\cref{sec:method_robust_deltas}) with CFG & - & - & 3.15 & - & - & 0.73 & 0.39 & \cmark & 23.0s [2.17it/s]\\
                \bottomrule
            \end{tabular}
            }
            \captionof{table}{Quantitative comparison with other control methods. We evaluate \textbf{(a)} subject-specificity of control in multi-subject settings, \textbf{(b)} disentangledness of attribute control \textit{v.s.}\ overall image changes, where we normalize the change metrics $\Delta \text{Id}$ and $\mathrm{LPIPS}$ by the attribute expression change $|\Delta \mathrm{CLIP}_\mathrm{Bi}|$, \textbf{(c)}~whether the method can be used for {fully}/uninterrupted continuous control from the original image, and \textbf{(d)} image generation speed (using an Nvidia A100 at batch size 1).}
            \label{tab:main_quantitative}
        \end{minipage}%
        \hspace{.02\textwidth}%
        \begin{minipage}{.39\textwidth}
            \vspace{4mm}
            \begin{minipage}{.9\linewidth}%
                \vspace{-5.6mm}
                \adjustbox{max width=\linewidth}{
                \begin{minipage}{1.71\actuallinewidth}
                    \input{figs/age_qualitative/figure}
                \end{minipage}
                }
                \vspace{-2.9mm}
            \end{minipage}
            \captionof{figure}{Qualitative comparison with other methods. (a) We continuously modulate the age of a person. (b) P2P~\citep{hertz2023prompttoprompt} and MasaCtrl~\citep{cao2023masactrl} do not offer full continuous control, first modulating to ``old'' or ``young'' and then optionally reweighting the adjective from there in the case of P2P. Unmodulated images are marked in \ourhighlightcolorstr.}
            \label{fig:age_qualitative}
        \end{minipage}
    \end{minipage}
    }
\end{figure*}

In this section, we comprehensively evaluate our proposed method. We conduct experiments by applying our semantic directions to both biasing the distribution of generated images and editing real images. We validate key properties such as subject-specificity, the disentanglement of edits, the fine-grainedness of control, and inference performance, and show that no current controlled generation method offers \textit{both} continuous \textit{and} subject-specific control simultaneously.

\subsection{Experimental Setup}
We evaluate our proposed method primarily on Stable Diffusion XL \citep{podell2024sdxl}, a widely used large-scale T2I diffusion model.
To test our method, we obtain a large variety of semantic directions for various attributes, primarily focused on humans, but also including vehicles and furniture. Detailed training procedures and parameters are in \cref{sec:app_training_details}.

\paragraph{Integration with other methods}
As our modulations augment the text prompt embedding input without adapting the model, they can directly be combined with many controlled generation and editing methods that utilize prompt changes for control, augmenting them with more fine-grained control. As part of our experiments, we demonstrate this integration with both Prompt-to-Prompt (P2P) \citep{hertz2023prompttoprompt} and AdapEdit \citep{ma2024adapedit}, where we simply replace their text modifications with our attribute modulations. Both methods improve consistency with an original generated image when changing the prompt. This combines the benefits of improved disentanglement and structure retainment of these methods with the more fine-grained control of our modulations.
We also combine our method with inversion using ReNoise~\citep{garibi2024renoise} to perform real image editing (see \cref{fig:real_image_editing}). Our combination with AdapEdit uses SD 1.5~\citep{ma2024adapedit}, as AdapEdit is not available for SDXL. Similarly, we use ReNoise with SDXL Turbo~\citep{sauer2023adversarial}.

\begin{figure}[t]
    \centering
    \vspace{-3mm}
    \begin{minipage}{.8\linewidth}%
        \adjustbox{max width=\linewidth}{
        \begin{minipage}{1.1\actuallinewidth}
            \input{figs/editing_qualitative/figure}
        \end{minipage}
        }
    \end{minipage}
    \vspace{-4mm}
    \caption{Real image editing: we apply our method to editing by inverting the image with ReNoise~\citep{garibi2024renoise} and regenerating the image with our modulations applied.}
    \label{fig:real_image_editing}
\end{figure}

\subsection{Attribute Control for Image Generation}
We evaluate our method’s ability to control attribute expression for specific target attributes $\attributei$ in different settings and compare it against other approaches both quantitatively and qualitatively. Full descriptions of our experimental setup and evaluation protocols are available in \cref{sec:app_experiment_evaluation_details}.

\paragraph{Subject-Specificity of Control}

To evaluate subject-specificity, we apply different attribute modulations to individual subjects within multi-subject-prompts. As shown in \cref{fig:attribute_delta_composability}b (see also \cref{sec:app_multi_subject_attribute_editing} for additional examples), our method can apply attribute modulations independently to each subject $\subjectj \in \subjects$ in multi-subject prompts $\prompt$, yielding fine-grained, disentangled control. This is despite training the directions $\pembdiff_{\attributei}$ only in a single-subject setting. We also find that our modulations enable an extensive coverage of the 2D attribute expression space when applied to multi-subject modulations, improving upon the coverage achieved by other methods (see \cref{fig:age_coverage}).

For a quantitative evaluation, we use two-subject prompts containing a 
target entity $\subject{\text{target}}$ and another $\subject{\text{other}}$ of the same category and measure the change induced by modulating an attribute of one subject relative to the other. Using detected bounding boxes, we calculate the change in CLIP score (a standard metric often used to quantify semantic control magnitudes \citep{gandikota2023sliders,ma2024adapedit}) 
for both $\subject{\text{target}}$ and the other subject $\subject{\text{other}}$ as:
\begin{multline}
\Delta \mathrm{CLIP} = 100 \cdot ( \mathrm{cossim}_\mathrm{CLIP}(\mathcal{I}_\mathrm{mod}, \prompt_\mathrm{edit})\\- \mathrm{cossim}_\mathrm{CLIP}(\mathcal{I}_\mathrm{orig}, \prompt_\mathrm{edit}))
\end{multline}
where $I_\mathrm{orig}$ and $I_\mathrm{mod}$ denote the original and edited images, respectively, and $P_\mathrm{edit}$ is the desired attribute edit prompt. The similarity $\mathrm{cossim}_\mathrm{CLIP}$ measures the alignment between the CLIP embeddings of the images and the attribute-edit prompts. 
From this, we compute the subject-specificity ratio by comparing the semantic $\Delta \text{CLIP}$ change for the target subject's image bbox $[\subject{\text{target}}]$ to the other subject $[\subject{\text{other}}]$. Formally, we define the subject-specificity metric as:
\begin{equation}\label{eq:specificity}
\text{Subject-Specificity} = \frac{|\Delta\mathrm{CLIP}(\mathcal{I}_{\mathrm{mod},[\subject{\text{target}}]})|}{|\Delta \mathrm{CLIP}(\mathcal{I}_{\mathrm{mod},[\subject{\text{other}}]})|}.
\end{equation}
As shown in our evaluation against state-of-the-art control methods in \cref{tab:main_quantitative}a, our method retains subject-specificity similar to adding adjectives to the prompt and Prompt-to-Prompt~\citep{hertz2023prompttoprompt}, allowing it to achieve fairly isolated changes in attribute expression. AdapEdit, which does not allow continuous modulations, performs substantially better. As AdapEdit uses text prompts to specify changes, we can combine it with our method (unlike other continuous modulation methods such as Concept Sliders, which can not be combined this way) to retain the superior subject-specificity, but also achieve continuous modulations.

\begin{figure}[tb]
    \centering
    \adjustbox{max width=\linewidth}{
        \begin{adjustbox}{clip, trim={3mm} {3mm} {1.8\columnwidth} {4.5mm}}
            \input{figs/coverage_eval.pgf}
        \end{adjustbox}
    }
    \adjustbox{max width=\linewidth}{
        \begin{adjustbox}{clip, trim={3mm} {4mm} {3.45\columnwidth} {4.5mm}}
            \input{figs/coverage_eval.pgf}
        \end{adjustbox}%
        \begin{adjustbox}{clip, trim={1.98\columnwidth} {4mm} {3mm} {4.5mm}}
            \input{figs/coverage_eval.pgf}
        \end{adjustbox}
    }
    \caption{We continuously modulate the target attribute for each of two subjects and estimate the individual attribute expression $\expr_{\subjectj}(\attributei)$ of the target attribute. Our modulations enable reaching a large range of attribute expression combinations, as they are both subject-specific and fully continuous. Other methods are limited in one of these aspects and thus do not allow full coverage. Samples with AdapEdit use SD 1.5, while the rest use SDXL.}
    \label{fig:age_coverage}
\end{figure}

\paragraph{Disentangledness of Control}

We also evaluate how disentangled the achieved semantic modulation is from both overall image changes and person identity changes (when applying modulations to people). We quantify overall perceptual image change using LPIPS \citep{lpips} and for identity similarity, we use the cosine similarity in the ReID embedding space, denoted as $\mathrm{cossim}_\mathrm{ReID}$, based on ArcFace embeddings \citep{deng2019arcface}. The identity change is computed as:
\begin{equation}\label{eq:delta_id}
   \Delta \mathrm{Id} = 1 - \mathrm{cossim}_\mathrm{ReID}(\mathcal{I}_\text{mod}, \mathcal{I}_\text{orig}),
\end{equation}
We show both results over the magnitude of the achieved semantic change in \cref{fig:continuous_modulation_image_change}, quantifying the semantic change as a bidirectional CLIP score change:
\begin{equation}\label{eq:delta_clip_bi}
    \Delta \mathrm{CLIP}_\mathrm{Bi} = \Delta \mathrm{CLIP}_{+} - \Delta \mathrm{CLIP}_{-},
\end{equation}
where $\Delta \mathrm{CLIP}_{+}$ uses a positive prompt (e.g., ``an old man") and $\Delta \mathrm{CLIP}_{-}$ uses a negative prompt (e.g., ``a young man").

This approach enables us to quantify both positive and negative changes in attribute expression faithfully.
We also consolidate these results into a single quantitative ratio each for image and person identity change in \cref{tab:main_quantitative}b. Compared to other methods, the attribute expression changes achieved with Attribute Control are well-disentangled from auxiliary image changes. When combined with AdapEdit, our method significantly outperforms all other approaches.

\begin{figure}[t]
    \centering
    \adjustbox{max width=1.1\columnwidth}{
        \input{figs/continuous_eval.pgf}
    }
    \caption{We measure the perceptual change in the image (LPIPS) and the person identity change ($\Delta\mathrm{Id}$) to the unmodified image while modulating the target attribute. Our modulations enable fully continuous and highly disentangled modulations, which is further improved by combining our method with others such as Prompt-to-Prompt or AdapEdit.}
    \label{fig:continuous_modulation_image_change}
\end{figure}

\begin{figure}[tb]
    \centering
    \adjustbox{max width=\linewidth}{
        \input{figs/compositional/figure}
    }
    \caption{\textbf{(a)} Multiple modulations can be composed simply by adding them. \textbf{(b)} Modulations can be applied to different subjects with different magnitudes. Unmodified images are marked \ourhighlightcolorstr.}
    \label{fig:attribute_delta_composability}
\end{figure}

\paragraph{Fine-Grainedness of Control}

We further demonstrate the fine-grained control capabilities of our method by showing smooth, gradual modifications in attribute expression across multiple target categories in \cref{fig:various_deltas}, qualitatively compared to other methods in \cref{fig:age_qualitative}, and quantitatively evaluated in \cref{fig:continuous_modulation_image_change}. Unlike other methods such as MasaCtrl~\citep{cao2023masactrl}, AdapEdit~\citep{ma2024adapedit}, or P2P \citep{hertz2023prompttoprompt}, which do not allow for fine-grained modulations, our approach enables continuous, well-disentangled modulation across a wide range of attribute expression $\expr(\attributei)$ similar to Concept Sliders \citep{gandikota2023sliders}, but while offering subject-specificity. This can also be seen in the multi-subject evaluation in \cref{fig:age_coverage}.

\begin{figure}[b]
    \centering
    \begin{minipage}{.9\linewidth}
        \adjustbox{max width=\linewidth}{
        \begin{minipage}{1.71\actuallinewidth}
            \input{figs/experiments_main_qualitative/single_attribute_reel_short}
        \end{minipage}
        }
        \vspace{-3mm}
    \end{minipage}
    \caption{Our modulations allow fine-grained control of attributes over many categories. Unmodulated images are marked \ourhighlightcolorstr. As changes are fine-grained and smooth, we recommend zooming in.}
    \label{fig:various_deltas}
\end{figure}

\paragraph{Ablation}

We also ablate over different variations of our method (see \cref{tab:main_quantitative}). We find that only applying the modulation after the first 20$\%$ of steps in the sampling process substantially improves the disentangledness of modulations. Furthermore, we find that our learning-based method for identifying modulation directions significantly improves upon the simple approach introduced in \cref{sec:method_naive_deltas}. Similarly, our learned directions are substantially more disentangled than just applying the $\pembdiff$ modulation they were trained on with Classifier-free Guidance (CFG)~\citep{ho2021classifierfree} and do not incur the substantial sampling cost overhead.

\begin{figure}[t]
    \centering
    \vspace{-.75mm}
    \newcommand{\partspacing}{1.5mm}
    \newcommand{\imgheight}{.14\linewidth}
    \begin{minipage}{\linewidth}
        \adjustbox{max width=\linewidth}{
        \begin{tikzpicture}
            \node[simple node image] (reel1) {\includegraphics[height=\imgheight]{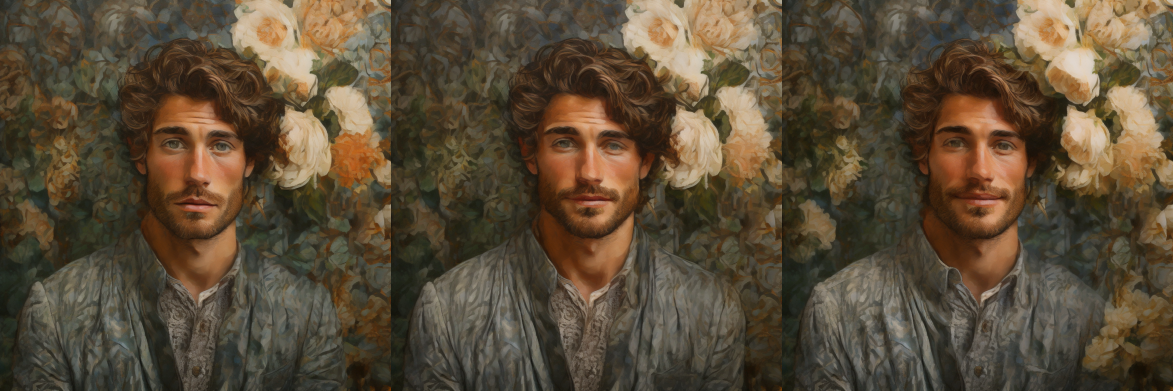}};
            \node[simple node image,right=\partspacing of reel1] (reel2) {\includegraphics[height=\imgheight]{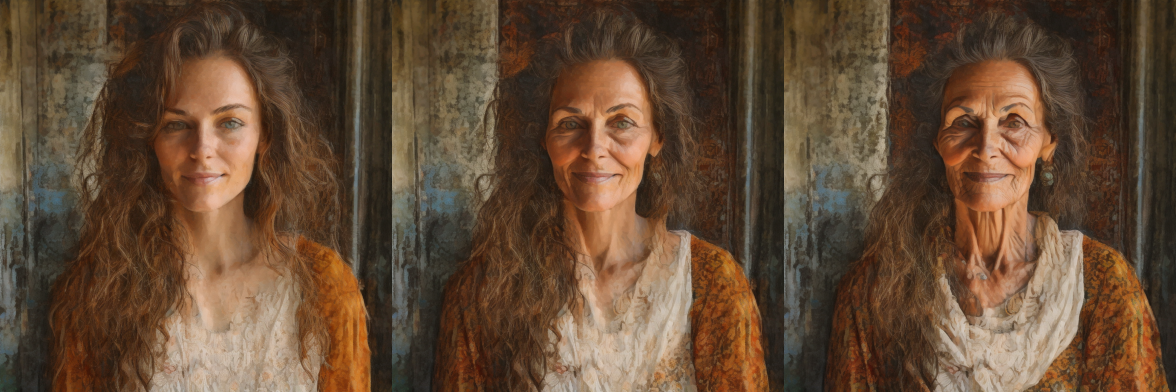}};
            \node[simple node image,right=\partspacing of reel2] (reel3) {\includegraphics[height=\imgheight]{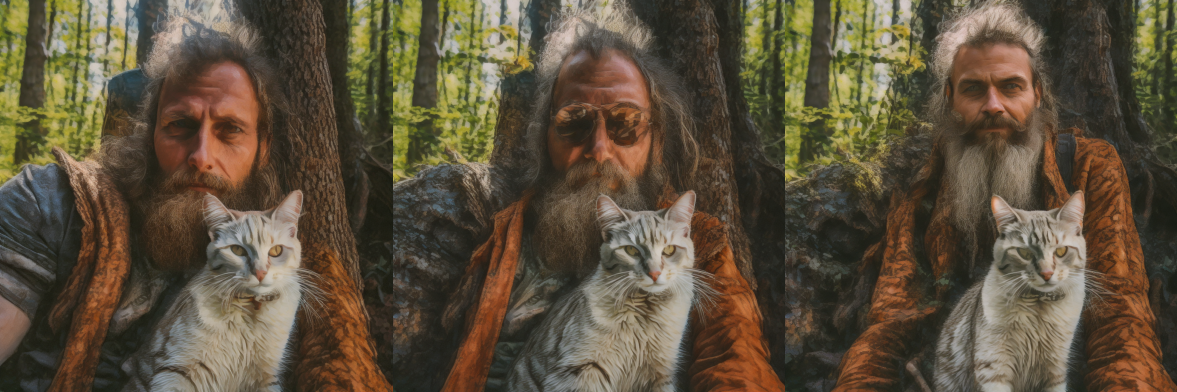}};
            
            \draw[standard double arrow] ($(reel1.west |- reel1.north) + (0, 1mm)$) -- node[above,yshift=-1mm] {Happiness} ($(reel1.east |- reel1.north) + (0, 1mm)$);
            \draw[standard arrow] ($(reel2.west |- reel2.north) + (0, 1mm)$) -- node[above,yshift=-1mm] {Age} ($(reel2.east |- reel2.north) + (0, 1mm)$);
            \draw[standard arrow] ($(reel3.west |- reel3.north) + (0, 1mm)$) -- node[above,yshift=-1mm] {Strangeness} ($(reel3.east |- reel3.north) + (0, 1mm)$);
            \node[draw=ourhighlightcolor,line width=0.04cm,below=0mm of reel1,minimum width=.143\linewidth,minimum height=.143\linewidth,yshift=.145\linewidth] {};
            \node[draw=ourhighlightcolor,line width=0.04cm,below=0mm of reel2,minimum width=.143\linewidth,minimum height=.143\linewidth,yshift=.145\linewidth,xshift=-.14\linewidth] {};
            \node[draw=ourhighlightcolor,line width=0.04cm,below=0mm of reel3,minimum width=.143\linewidth,minimum height=.143\linewidth,yshift=.145\linewidth,xshift=-.14\linewidth] {};
        \end{tikzpicture}
        \vspace{-3mm}
        }
    \end{minipage}
    \caption{Attribute Control with PixArt-$\alpha$ (T5-XXL text encoder).}
    \vspace{-1.5mm}
    \label{fig:pixart_alpha}
\end{figure}

\paragraph{Generalization}\label{sec:exp_generalization}
We further investigate the generalizability of our method.
Generally, any learned modulation direction $\pembdiff_{\attributei}$ will have only been trained on a closed set of nouns describing the target subject $\subject{}$. To verify that they generalize beyond this set, we apply directions that have been trained on a very small set of generic nouns (e.g., ``person'', ``woman'', and ``man'' for people) to more specific nouns (see \cref{sec:app_subject_noun_transferability}). We find that our directions generalize to this setting as expected.
We also find that our learned modulation directions $\pembdiff_{\attributei}$ can generalize to other models that use the same text encoders in a \textit{zero-shot} manner. By learning a direction on one model, in this case, SDXL \citep{podell2024sdxl}, we can directly transfer it to models that use the same text encoders (see \cref{fig:zero_shot_transfer}), such as SDXL Turbo \citep{sauer2023adversarial}, or a subset of them, as with SD 1.5 \citep{rombach2022high} or the image+depth model LDM3D~\citep{stan2023ldm3d}. Our learned directions even generalize to non-diffusion models such as aMUSEd \citep{patil2024amused}.
Our method can also generalize to models without CLIP text encoders, such as PixArt-$\alpha$~\cite{chen2023pixart}, which uses T5-XXL~\cite{raffel2020exploring} as shown in \cref{fig:pixart_alpha}.

\section{Conclusion}
This work uncovers the powerful capabilities of the tokenwise CLIP \cite{clip} text embedding for exerting control over the image generation process in T2I diffusion models. Instead of just acting as a discrete space of embeddings of words, we find that diffusion models are capable of interpreting local deviations in the tokenwise CLIP text embedding space in semantically meaningful ways. We use this insight to augment the typically rather coarse prompt with fine-grained, continuous control over the attribute expression of specific subjects by identifying semantic directions that correspond to specific attributes. Since we only modify the tokenwise CLIP text embedding along pre-identified directions, we enable more fine-grained manipulation at no additional cost in the generation process.

\begin{figure}[t]
    \centering
    \begin{minipage}{.65\linewidth}\hspace{-3mm}
        \adjustbox{max width=\linewidth}{
        \begin{minipage}{1.23\actuallinewidth}
            \input{figs/zero_shot_amused/figure_horizontal}
        \end{minipage}
        }
    \end{minipage}
    \vspace{-1mm}
    \caption{Zero-shot transfer: our modulations can be learned on one model (SDXL) and transferred to others (including non-diffusion models) without re-training. This also allows us to combine them with methods for other models, such as AdapEdit~\citep{ma2024adapedit} on SD 1.5, which does not offer continuous subject-specific modulations by itself. Unmodulated images are marked in \ourhighlightcolorstr.}
    \label{fig:zero_shot_transfer}
\end{figure}

\paragraph{Limitations and Future Work}

This work is a step towards revealing the hidden
capabilities of the text embedding input to common large-scale
diffusion models and making them usable in straightforward ways. While our approach works for different off-the-shelf models without modifying them, it is also inherently limited by their capabilities. Specifically, our method inherits the limitation that diffusion models sometimes mix up attributes between different subjects. Complementary methods \citep{chefer2023attendexcite,rassin2023linguistic,xiao2024fastcomposer} reduce these problems substantially, and future work could investigate their combination with our method in depth. Our approach also uses linear modulations along semantic directions in CLIP's tokenwise embedding space. In GANs, where similar linear modulations are often used, previous works~\citep{balakrishnan2022rayleigh} found that more disentangled changes can be achieved using nonlinear modulations. The tokenwise CLIP text embedding space might share this property and could benefit from applying similar strategies to further improve disentanglement.

\section*{Acknowledgement}
We thank Timy Phan for proofreading and feedback.
This project has been supported by the German Federal Ministry for Economic Affairs and Climate Action within the project ``NXT GEN AI METHODS – Generative Methoden für Perzeption, Prädiktion und Planung'', the project ``GeniusRobot'' (01IS24083), funded by the Federal Ministry of Education and Research (BMBF), the bidt project KLIMA-MEMES, and the German Research Foundation (DFG) project 421703927.
The authors acknowledge the Gauss Center for Supercomputing for providing compute through the NIC on JUWELS at JSC and the HPC resources supplied by the Erlangen National High Performance Computing Center (NHR@FAU funded by DFG project 440719683) under the NHR project JA-22883.

{
    \small
    \bibliographystyle{ieeenat_fullname}
    \bibliography{bib}

\begin{thebibliography}{64}
\providecommand{\natexlab}[1]{#1}
\providecommand{\url}[1]{\texttt{#1}}
\expandafter\ifx\csname urlstyle\endcsname\relax
  \providecommand{\doi}[1]{doi: #1}\else
  \providecommand{\doi}{doi: \begingroup \urlstyle{rm}\Url}\fi

\bibitem[Abdal et~al.(2019)Abdal, Qin, and Wonka]{abdal2019image2styleganembedimagesstylegan}
Rameen Abdal, Yipeng Qin, and Peter Wonka.
\newblock Image2stylegan: How to embed images into the stylegan latent space?
\newblock In \emph{Proceedings of the IEEE/CVF International Conference on Computer Vision (ICCV)}, 2019.

\bibitem[Abdal et~al.(2022)Abdal, Zhu, Femiani, Mitra, and Wonka]{abdal2022clip2stylegan}
Rameen Abdal, Peihao Zhu, John Femiani, Niloy Mitra, and Peter Wonka.
\newblock Clip2stylegan: Unsupervised extraction of stylegan edit directions.
\newblock 2022.

\bibitem[Balakrishnan et~al.(2022)Balakrishnan, Gadde, Martinez, and Perona]{balakrishnan2022rayleigh}
Guha Balakrishnan, Raghudeep Gadde, Aleix Martinez, and Pietro Perona.
\newblock Rayleigh eigendirections (reds): Nonlinear gan latent space traversals for multidimensional features.
\newblock In \emph{European Conference on Computer Vision}, pages 510--526. Springer, 2022.

\bibitem[Betker et~al.(2023)Betker, Goh, Jing, Brooks, Wang, Li, Ouyang, Zhuang, Lee, Guo, et~al.]{betker2023dalle3}
James Betker, Gabriel Goh, Li Jing, Tim Brooks, Jianfeng Wang, Linjie Li, Long Ouyang, Juntang Zhuang, Joyce Lee, Yufei Guo, et~al.
\newblock Improving image generation with better captions, 2023.

\bibitem[Brooks et~al.(2023)Brooks, Holynski, and Efros]{brooks2023instructpix2pix}
Tim Brooks, Aleksander Holynski, and Alexei~A Efros.
\newblock Instructpix2pix: Learning to follow image editing instructions.
\newblock In \emph{Proceedings of the IEEE/CVF Conference on Computer Vision and Pattern Recognition}, 2023.

\bibitem[Cao et~al.(2023)Cao, Wang, Qi, Shan, Qie, and Zheng]{cao2023masactrl}
Mingdeng Cao, Xintao Wang, Zhongang Qi, Ying Shan, Xiaohu Qie, and Yinqiang Zheng.
\newblock Masactrl: Tuning-free mutual self-attention control for consistent image synthesis and editing.
\newblock In \emph{Proceedings of the IEEE/CVF International Conference on Computer Vision}, pages 22560--22570, 2023.

\bibitem[Chefer et~al.(2023)Chefer, Alaluf, Vinker, Wolf, and Cohen-Or]{chefer2023attendexcite}
Hila Chefer, Yuval Alaluf, Yael Vinker, Lior Wolf, and Daniel Cohen-Or.
\newblock Attend-and-excite: Attention-based semantic guidance for text-to-image diffusion models.
\newblock \emph{ACM Trans. Graph.}, 42\penalty0 (4), 2023.

\bibitem[Chefer et~al.(2024)Chefer, Lang, Geva, Polosukhin, Shocher, michal Irani, Mosseri, and Wolf]{chefer2024hiddenlanguage}
Hila Chefer, Oran Lang, Mor Geva, Volodymyr Polosukhin, Assaf Shocher, michal Irani, Inbar Mosseri, and Lior Wolf.
\newblock The hidden language of diffusion models.
\newblock In \emph{The Twelfth International Conference on Learning Representations}, 2024.

\bibitem[Chen et~al.(2023)Chen, Yu, Ge, Yao, Xie, Wu, Wang, Kwok, Luo, Lu, et~al.]{chen2023pixart}
Junsong Chen, Jincheng Yu, Chongjian Ge, Lewei Yao, Enze Xie, Yue Wu, Zhongdao Wang, James Kwok, Ping Luo, Huchuan Lu, et~al.
\newblock Pixart-$\alpha$: Fast training of diffusion transformer for photorealistic text-to-image synthesis.
\newblock \emph{arXiv preprint arXiv:2310.00426}, 2023.

\bibitem[Darcet et~al.(2024)Darcet, Oquab, Mairal, and Bojanowski]{registers}
Timoth{\'e}e Darcet, Maxime Oquab, Julien Mairal, and Piotr Bojanowski.
\newblock Vision transformers need registers.
\newblock In \emph{The Twelfth International Conference on Learning Representations}, 2024.

\bibitem[Deng et~al.(2019)Deng, Guo, Xue, and Zafeiriou]{deng2019arcface}
Jiankang Deng, Jia Guo, Niannan Xue, and Stefanos Zafeiriou.
\newblock Arcface: Additive angular margin loss for deep face recognition.
\newblock In \emph{Proceedings of the IEEE/CVF Conference on Computer Vision and Pattern Recognition}, 2019.

\bibitem[Deutch et~al.(2024)Deutch, Gal, Garibi, Patashnik, and Cohen-Or]{deutch2024turboedit}
Gilad Deutch, Rinon Gal, Daniel Garibi, Or Patashnik, and Daniel Cohen-Or.
\newblock Turboedit: Text-based image editing using few-step diffusion models.
\newblock In \emph{SIGGRAPH Asia 2024 Conference Papers}, pages 1--12, 2024.

\bibitem[Dhariwal and Nichol(2021)]{dhariwal2021adm}
Prafulla Dhariwal and Alexander Nichol.
\newblock Diffusion models beat gans on image synthesis.
\newblock In \emph{Advances in Neural Information Processing Systems}. Curran Associates, Inc., 2021.

\bibitem[Gandikota et~al.(2024)Gandikota, Materzy\'nska, Zhou, Torralba, and Bau]{gandikota2023sliders}
Rohit Gandikota, Joanna Materzy\'nska, Tingrui Zhou, Antonio Torralba, and David Bau.
\newblock Concept sliders: Lora adaptors for precise control in diffusion models.
\newblock In \emph{European Conference on Computer Vision}, 2024.

\bibitem[Garibi et~al.(2024)Garibi, Patashnik, Voynov, Averbuch-Elor, and Cohen-Or]{garibi2024renoise}
Daniel Garibi, Or Patashnik, Andrey Voynov, Hadar Averbuch-Elor, and Daniel Cohen-Or.
\newblock Renoise: Real image inversion through iterative noising.
\newblock In \emph{European Conference on Computer Vision}, 2024.

\bibitem[Goodfellow et~al.(2014)Goodfellow, Pouget-Abadie, Mirza, Xu, Warde-Farley, Ozair, Courville, and Bengio]{goodfellow2014generative}
Ian Goodfellow, Jean Pouget-Abadie, Mehdi Mirza, Bing Xu, David Warde-Farley, Sherjil Ozair, Aaron Courville, and Yoshua Bengio.
\newblock Generative adversarial nets.
\newblock \emph{Advances in neural information processing systems}, 27, 2014.

\bibitem[Hertz et~al.(2023)Hertz, Mokady, Tenenbaum, Aberman, Pritch, and Cohen-or]{hertz2023prompttoprompt}
Amir Hertz, Ron Mokady, Jay Tenenbaum, Kfir Aberman, Yael Pritch, and Daniel Cohen-or.
\newblock Prompt-to-prompt image editing with cross-attention control.
\newblock In \emph{The Eleventh International Conference on Learning Representations}, 2023.

\bibitem[Ho and Salimans(2021)]{ho2021classifierfree}
Jonathan Ho and Tim Salimans.
\newblock Classifier-free diffusion guidance.
\newblock In \emph{NeurIPS 2021 Workshop on Deep Generative Models and Downstream Applications}, 2021.

\bibitem[Ho et~al.(2020)Ho, Jain, and Abbeel]{ho2020denoising}
Jonathan Ho, Ajay Jain, and Pieter Abbeel.
\newblock Denoising diffusion probabilistic models.
\newblock \emph{Advances in neural information processing systems}, 33:\penalty0 6840--6851, 2020.

\bibitem[Hu et~al.(2022)Hu, yelong shen, Wallis, Allen-Zhu, Li, Wang, Wang, and Chen]{hu2021lora}
Edward~J Hu, yelong shen, Phillip Wallis, Zeyuan Allen-Zhu, Yuanzhi Li, Shean Wang, Lu Wang, and Weizhu Chen.
\newblock Lo{RA}: Low-rank adaptation of large language models.
\newblock In \emph{International Conference on Learning Representations}, 2022.

\bibitem[Hu et~al.(2024)Hu, Zhang, Tang, Mettes, Zhao, and Snoek]{hu2024latentspaceediting}
Vincent~Tao Hu, Wei Zhang, Meng Tang, Pascal Mettes, Deli Zhao, and Cees Snoek.
\newblock Latent space editing in transformer-based flow matching.
\newblock \emph{Proceedings of the AAAI Conference on Artificial Intelligence}, 38\penalty0 (3):\penalty0 2247--2255, 2024.

\bibitem[Huberman-Spiegelglas et~al.(2024)Huberman-Spiegelglas, Kulikov, and Michaeli]{huberman2024edit}
Inbar Huberman-Spiegelglas, Vladimir Kulikov, and Tomer Michaeli.
\newblock An edit friendly ddpm noise space: Inversion and manipulations.
\newblock In \emph{Proceedings of the IEEE/CVF Conference on Computer Vision and Pattern Recognition}, pages 12469--12478, 2024.

\bibitem[Ilharco et~al.(2021)Ilharco, Wortsman, Wightman, Gordon, Carlini, Taori, Dave, Shankar, Namkoong, Miller, Hajishirzi, Farhadi, and Schmidt]{openclip_software}
Gabriel Ilharco, Mitchell Wortsman, Ross Wightman, Cade Gordon, Nicholas Carlini, Rohan Taori, Achal Dave, Vaishaal Shankar, Hongseok Namkoong, John Miller, Hannaneh Hajishirzi, Ali Farhadi, and Ludwig Schmidt.
\newblock Openclip, 2021.

\bibitem[Imagen-Team(2024)]{imagenteamgoogle2024imagen3}
Imagen-Team.
\newblock Imagen 3, 2024.

\bibitem[Karras et~al.(2019)Karras, Laine, and Aila]{karras2019style}
Tero Karras, Samuli Laine, and Timo Aila.
\newblock A style-based generator architecture for generative adversarial networks.
\newblock In \emph{Proceedings of the IEEE/CVF conference on computer vision and pattern recognition}, pages 4401--4410, 2019.

\bibitem[Kim et~al.(2022)Kim, Kwon, and Ye]{diffusionclip}
Gwanghyun Kim, Taesung Kwon, and Jong~Chul Ye.
\newblock Diffusionclip: Text-guided diffusion models for robust image manipulation.
\newblock In \emph{Proceedings of the IEEE/CVF Conference on Computer Vision and Pattern Recognition (CVPR)}, pages 2426--2435, 2022.

\bibitem[Kwon et~al.(2023)Kwon, Jeong, and Uh]{kwon2023diffusion}
Mingi Kwon, Jaeseok Jeong, and Youngjung Uh.
\newblock Diffusion models already have a semantic latent space.
\newblock In \emph{The Eleventh International Conference on Learning Representations}, 2023.

\bibitem[Li et~al.(2024{\natexlab{a}})Li, Shen, Torr, Tresp, and Gu]{li2024self}
Hang Li, Chengzhi Shen, Philip Torr, Volker Tresp, and Jindong Gu.
\newblock Self-discovering interpretable diffusion latent directions for responsible text-to-image generation.
\newblock In \emph{Proceedings of the IEEE/CVF Conference on Computer Vision and Pattern Recognition}, pages 12006--12016, 2024{\natexlab{a}}.

\bibitem[Li et~al.(2024{\natexlab{b}})Li, van~de Weijer, taihang Hu, Khan, Hou, Wang, and jian Yang]{li2024getwhatyouwant}
Senmao Li, Joost van~de Weijer, taihang Hu, Fahad Khan, Qibin Hou, Yaxing Wang, and jian Yang.
\newblock Get what you want, not what you don't: Image content suppression for text-to-image diffusion models.
\newblock In \emph{The Twelfth International Conference on Learning Representations}, 2024{\natexlab{b}}.

\bibitem[Li et~al.(2024{\natexlab{c}})Li, Zeng, Feng, Gao, Liu, Liu, Li, Tang, Hu, Liu, et~al.]{li2024zone}
Shanglin Li, Bohan Zeng, Yutang Feng, Sicheng Gao, Xiuhui Liu, Jiaming Liu, Lin Li, Xu Tang, Yao Hu, Jianzhuang Liu, et~al.
\newblock Zone: Zero-shot instruction-guided local editing.
\newblock In \emph{Proceedings of the IEEE/CVF Conference on Computer Vision and Pattern Recognition}, pages 6254--6263, 2024{\natexlab{c}}.

\bibitem[Loshchilov and Hutter(2019)]{loshchilov2018decoupled}
Ilya Loshchilov and Frank Hutter.
\newblock Decoupled weight decay regularization.
\newblock In \emph{International Conference on Learning Representations}, 2019.

\bibitem[Ma et~al.(2024)Ma, Jia, and Zhou]{ma2024adapedit}
Zhiyuan Ma, Guoli Jia, and Bowen Zhou.
\newblock Adapedit: Spatio-temporal guided adaptive editing algorithm for text-based continuity-sensitive image editing.
\newblock \emph{Proceedings of the AAAI Conference on Artificial Intelligence}, 38\penalty0 (5):\penalty0 4154--4161, 2024.

\bibitem[Mao et~al.(2023)Mao, Chen, Gu, Fang, and Shou]{mao2023mag}
Qi Mao, Lan Chen, Yuchao Gu, Zhen Fang, and Mike~Zheng Shou.
\newblock Mag-edit: Localized image editing in complex scenarios via mask-based attention-adjusted guidance.
\newblock \emph{arXiv}, 2023.

\bibitem[Meng et~al.(2022)Meng, He, Song, Song, Wu, Zhu, and Ermon]{meng2021sdedit}
Chenlin Meng, Yutong He, Yang Song, Jiaming Song, Jiajun Wu, Jun-Yan Zhu, and Stefano Ermon.
\newblock {SDE}dit: Guided image synthesis and editing with stochastic differential equations.
\newblock In \emph{International Conference on Learning Representations}, 2022.

\bibitem[Mirzaei et~al.(2024)Mirzaei, Aumentado-Armstrong, Brubaker, Kelly, Levinshtein, Derpanis, and Gilitschenski]{mirzaei2024watch}
Ashkan Mirzaei, Tristan Aumentado-Armstrong, Marcus~A Brubaker, Jonathan Kelly, Alex Levinshtein, Konstantinos~G Derpanis, and Igor Gilitschenski.
\newblock Watch your steps: Local image and scene editing by text instructions.
\newblock In \emph{European Conference on Computer Vision}, pages 111--129. Springer, 2024.

\bibitem[Mokady et~al.(2023)Mokady, Hertz, Aberman, Pritch, and Cohen-Or]{mokady2023nulltextinversion}
Ron Mokady, Amir Hertz, Kfir Aberman, Yael Pritch, and Daniel Cohen-Or.
\newblock Null-text inversion for editing real images using guided diffusion models.
\newblock In \emph{Proceedings of the IEEE/CVF Conference on Computer Vision and Pattern Recognition (CVPR)}, pages 6038--6047, 2023.

\bibitem[Oquab et~al.(2024)Oquab, Darcet, Moutakanni, Vo, Szafraniec, Khalidov, Fernandez, HAZIZA, Massa, El-Nouby, Assran, Ballas, Galuba, Howes, Huang, Li, Misra, Rabbat, Sharma, Synnaeve, Xu, Jegou, Mairal, Labatut, Joulin, and Bojanowski]{dinov2}
Maxime Oquab, Timoth{\'e}e Darcet, Th{\'e}o Moutakanni, Huy~V. Vo, Marc Szafraniec, Vasil Khalidov, Pierre Fernandez, Daniel HAZIZA, Francisco Massa, Alaaeldin El-Nouby, Mido Assran, Nicolas Ballas, Wojciech Galuba, Russell Howes, Po-Yao Huang, Shang-Wen Li, Ishan Misra, Michael Rabbat, Vasu Sharma, Gabriel Synnaeve, Hu Xu, Herve Jegou, Julien Mairal, Patrick Labatut, Armand Joulin, and Piotr Bojanowski.
\newblock {DINO}v2: Learning robust visual features without supervision.
\newblock \emph{Transactions on Machine Learning Research}, 2024.

\bibitem[Patashnik et~al.(2021)Patashnik, Wu, Shechtman, Cohen-Or, and Lischinski]{patashnik2021styleclip}
Or Patashnik, Zongze Wu, Eli Shechtman, Daniel Cohen-Or, and Dani Lischinski.
\newblock Styleclip: Text-driven manipulation of stylegan imagery.
\newblock In \emph{Proceedings of the IEEE/CVF International Conference on Computer Vision (ICCV)}, pages 2085--2094, 2021.

\bibitem[Patil et~al.(2024)Patil, Berman, Rombach, and von Platen]{patil2024amused}
Suraj Patil, William Berman, Robin Rombach, and Patrick von Platen.
\newblock amused: An open muse reproduction.
\newblock \emph{arXiv}, 2024.

\bibitem[Podell et~al.(2024)Podell, English, Lacey, Blattmann, Dockhorn, M{\"u}ller, Penna, and Rombach]{podell2024sdxl}
Dustin Podell, Zion English, Kyle Lacey, Andreas Blattmann, Tim Dockhorn, Jonas M{\"u}ller, Joe Penna, and Robin Rombach.
\newblock {SDXL}: Improving latent diffusion models for high-resolution image synthesis.
\newblock In \emph{The Twelfth International Conference on Learning Representations}, 2024.

\bibitem[Radford et~al.(2016)Radford, Metz, and Chintala]{radford2015dcgan}
Alec Radford, Luke Metz, and Soumith Chintala.
\newblock Unsupervised representation learning with deep convolutional generative adversarial networks.
\newblock In \emph{4th International Conference on Learning Representations, {ICLR} 2016, San Juan, Puerto Rico, May 2-4, 2016, Conference Track Proceedings}, 2016.

\bibitem[Radford et~al.(2021)Radford, Kim, Hallacy, Ramesh, Goh, Agarwal, Sastry, Askell, Mishkin, Clark, et~al.]{clip}
Alec Radford, Jong~Wook Kim, Chris Hallacy, Aditya Ramesh, Gabriel Goh, Sandhini Agarwal, Girish Sastry, Amanda Askell, Pamela Mishkin, Jack Clark, et~al.
\newblock Learning transferable visual models from natural language supervision.
\newblock In \emph{International conference on machine learning}, pages 8748--8763. PMLR, 2021.

\bibitem[Raffel et~al.(2020)Raffel, Shazeer, Roberts, Lee, Narang, Matena, Zhou, Li, and Liu]{raffel2020exploring}
Colin Raffel, Noam Shazeer, Adam Roberts, Katherine Lee, Sharan Narang, Michael Matena, Yanqi Zhou, Wei Li, and Peter~J Liu.
\newblock Exploring the limits of transfer learning with a unified text-to-text transformer.
\newblock \emph{Journal of machine learning research}, 21\penalty0 (140):\penalty0 1--67, 2020.

\bibitem[Ramesh et~al.(2022)Ramesh, Dhariwal, Nichol, Chu, and Chen]{ramesh2022hierarchicaltextconditionalimagegeneration}
Aditya Ramesh, Prafulla Dhariwal, Alex Nichol, Casey Chu, and Mark Chen.
\newblock Hierarchical text-conditional image generation with clip latents, 2022.

\bibitem[Rassin et~al.(2023)Rassin, Hirsch, Glickman, Ravfogel, Goldberg, and Chechik]{rassin2023linguistic}
Royi Rassin, Eran Hirsch, Daniel Glickman, Shauli Ravfogel, Yoav Goldberg, and Gal Chechik.
\newblock Linguistic binding in diffusion models: Enhancing attribute correspondence through attention map alignment.
\newblock In \emph{Thirty-seventh Conference on Neural Information Processing Systems}, 2023.

\bibitem[Rombach et~al.(2022)Rombach, Blattmann, Lorenz, Esser, and Ommer]{rombach2022high}
Robin Rombach, Andreas Blattmann, Dominik Lorenz, Patrick Esser, and Bj{\"o}rn Ommer.
\newblock High-resolution image synthesis with latent diffusion models.
\newblock In \emph{Proceedings of the IEEE/CVF conference on computer vision and pattern recognition}, pages 10684--10695, 2022.

\bibitem[Sauer et~al.(2023)Sauer, Lorenz, Blattmann, and Rombach]{sauer2023adversarial}
Axel Sauer, Dominik Lorenz, Andreas Blattmann, and Robin Rombach.
\newblock Adversarial diffusion distillation.
\newblock \emph{arXiv preprint arXiv:2311.17042}, 2023.

\bibitem[Shen et~al.(2020)Shen, Yang, Tang, and Zhou]{shen2020interfacegan}
Yujun Shen, Ceyuan Yang, Xiaoou Tang, and Bolei Zhou.
\newblock Interfacegan: Interpreting the disentangled face representation learned by gans.
\newblock \emph{IEEE transactions on pattern analysis and machine intelligence}, 44\penalty0 (4):\penalty0 2004--2018, 2020.

\bibitem[Simsar et~al.(2023)Simsar, Tonioni, Xian, Hofmann, and Tombari]{simsar2023lime}
Enis Simsar, Alessio Tonioni, Yongqin Xian, Thomas Hofmann, and Federico Tombari.
\newblock Lime: Localized image editing via attention regularization in diffusion models.
\newblock \emph{arXiv}, 2023.

\bibitem[Song et~al.(2021)Song, Meng, and Ermon]{song2021ddim}
Jiaming Song, Chenlin Meng, and Stefano Ermon.
\newblock Denoising diffusion implicit models.
\newblock In \emph{International Conference on Learning Representations}, 2021.

\bibitem[Stan et~al.(2023)Stan, Wofk, Fox, Redden, Saxton, Yu, Aflalo, Tseng, Nonato, Muller, and Lal]{stan2023ldm3d}
Gabriela Ben~Melech Stan, Diana Wofk, Scottie Fox, Alex Redden, Will Saxton, Jean Yu, Estelle Aflalo, Shao-Yen Tseng, Fabio Nonato, Matthias Muller, and Vasudev Lal.
\newblock Ldm3d: Latent diffusion model for 3d.
\newblock In \emph{3DMV: Learning 3D with Multi-View Supervision (CVPRW’23)}, 2023.

\bibitem[Tsaban and Passos(2023)]{tsaban2023ledits}
Linoy Tsaban and Apolin{\'a}rio Passos.
\newblock Ledits: Real image editing with ddpm inversion and semantic guidance.
\newblock \emph{arXiv preprint arXiv:2307.00522}, 2023.

\bibitem[Wang et~al.(2023)Wang, Li, Lin, Lv, Schwing, and Ji]{wang2022learningdecomposevisualfeatures}
Feng Wang, Manling Li, Xudong Lin, Hairong Lv, Alex Schwing, and Heng Ji.
\newblock Learning to decompose visual features with latent textual prompts.
\newblock In \emph{The Eleventh International Conference on Learning Representations}, 2023.

\bibitem[Wang et~al.(2024)Wang, Gui, Negrea, and Veitch]{wang2024concept}
Zihao Wang, Lin Gui, Jeffrey Negrea, and Victor Veitch.
\newblock Concept algebra for (score-based) text-controlled generative models.
\newblock \emph{Advances in Neural Information Processing Systems}, 36, 2024.

\bibitem[Wu et~al.(2024{\natexlab{a}})Wu, Yang, and Wang]{wu2024relation}
Yinwei Wu, Xingyi Yang, and Xinchao Wang.
\newblock Relation rectification in diffusion model.
\newblock In \emph{Proceedings of the IEEE/CVF Conference on Computer Vision and Pattern Recognition}, pages 7685--7694, 2024{\natexlab{a}}.

\bibitem[Wu et~al.(2024{\natexlab{b}})Wu, Kolkin, Brandt, Zhang, and Shechtman]{wu2024turboedit}
Zongze Wu, Nicholas Kolkin, Jonathan Brandt, Richard Zhang, and Eli Shechtman.
\newblock Turboedit: Instant text-based image editing.
\newblock In \emph{European Conference on Computer Vision}, pages 365--381. Springer, 2024{\natexlab{b}}.

\bibitem[Xia et~al.(2021)Xia, Yang, Xue, and Wu]{xia2021tedigan}
Weihao Xia, Yujiu Yang, Jing-Hao Xue, and Baoyuan Wu.
\newblock Tedigan: Text-guided diverse face image generation and manipulation.
\newblock In \emph{Proceedings of the IEEE/CVF conference on computer vision and pattern recognition}, pages 2256--2265, 2021.

\bibitem[Xiao et~al.(2024)Xiao, Yin, Freeman, Durand, and Han]{xiao2024fastcomposer}
Guangxuan Xiao, Tianwei Yin, William~T Freeman, Fr{\'e}do Durand, and Song Han.
\newblock Fastcomposer: Tuning-free multi-subject image generation with localized attention.
\newblock \emph{International Journal of Computer Vision}, pages 1--20, 2024.

\bibitem[Yesiltepe et~al.(2024)Yesiltepe, Akdemir, and Yanardag]{yesiltepe2024mist}
Hidir Yesiltepe, Kiymet Akdemir, and Pinar Yanardag.
\newblock Mist: Mitigating intersectional bias with disentangled cross-attention editing in text-to-image diffusion models.
\newblock \emph{arXiv preprint arXiv:2403.19738}, 2024.

\bibitem[Zhang et~al.(2018)Zhang, Isola, Efros, Shechtman, and Wang]{lpips}
Richard Zhang, Phillip Isola, Alexei~A Efros, Eli Shechtman, and Oliver Wang.
\newblock The unreasonable effectiveness of deep features as a perceptual metric.
\newblock In \emph{CVPR}, 2018.

\bibitem[Zhang et~al.(2024)Zhang, Yang, Feng, Qin, Chen, Yu, Chen, Wang, Savarese, Ermon, et~al.]{zhang2023hive}
Shu Zhang, Xinyi Yang, Yihao Feng, Can Qin, Chia-Chih Chen, Ning Yu, Zeyuan Chen, Huan Wang, Silvio Savarese, Stefano Ermon, et~al.
\newblock Hive: Harnessing human feedback for instructional visual editing.
\newblock In \emph{Proceedings of the IEEE/CVF Conference on Computer Vision and Pattern Recognition}, pages 9026--9036, 2024.

\bibitem[Zhu et~al.(2020)Zhu, Shen, Zhao, and Zhou]{zhu2020indomainganinversionreal}
Jiapeng Zhu, Yujun Shen, Deli Zhao, and Bolei Zhou.
\newblock In-domain gan inversion for real image editing.
\newblock In \emph{European conference on computer vision}, pages 592--608. Springer, 2020.

\bibitem[Zhu et~al.(2023)Zhu, Wu, Deng, Russakovsky, and Yan]{zhu2023boundaryguided}
Ye Zhu, Yu Wu, Zhiwei Deng, Olga Russakovsky, and Yan Yan.
\newblock Boundary guided learning-free semantic control with diffusion models.
\newblock In \emph{Advances in Neural Information Processing Systems}, pages 78319--78346. Curran Associates, Inc., 2023.

\bibitem[Zhuang et~al.(2024)Zhuang, Hu, and Gao]{zhuang2024magnet}
Chenyi Zhuang, Ying Hu, and Pan Gao.
\newblock Magnet: We never know how text-to-image diffusion models work, until we learn how vision-language models function.
\newblock In \emph{The Thirty-eighth Annual Conference on Neural Information Processing Systems}, 2024.

\end{thebibliography}
}

\clearpage
\onecolumn
\raggedbottom

\setcounter{figure}{0}
\setcounter{table}{0}
\setcounter{section}{0}
\renewcommand\thefigure{\Alph{figure}}  
\renewcommand\thetable{\Alph{table}}  
\renewcommand\thesection{\Alph{section}}

\section{Additional Results}

\subsection{Additional Ablations}\label{sec:additional_ablations}
We show quantitative results for requested additional ablations in \cref{tab:additional_ablations}. Specifically, we investigate optimizing a variant of our training objective where the main change is omitting the attribute scale $\attributescalei$ variation. This performs substantially worse than the full version of our objective. We also evaluate directly taking the CLIP embedding of the target attribute -- either its general embedding as represented by the EOS token, or the relevant subject token. Both versions are similarly disentangled as our CLIP difference method, but substantially underperform compared to it in subject-specificity.

\begin{table}[H]
    \centering
    \adjustbox{max width=\linewidth}{
    \begin{tabular}{lHH@{}c@{}HHccc@{}r}
        & & & \multicolumn{1}{c}{(a) Subject-Specificity} & & & \multicolumn{2}{c}{(b) Disentangledness} & (c) & (d) Performance\\
        \midrule
        \textbf{Method} & & & \textbf{Subject-Specificity} $\uparrow$ & & &
        $\Delta\mathbf{Id}$ ${\downarrow}$ & $\mathbf{LPIPS}$ ${\downarrow}$ &
        \textbf{Continuous} & \textbf{Time} $\downarrow$\\
        \midrule
        Ours & \textbf{0.521} & {0.629} & \underline{3.35} & 0.072 & {0.019} & \textbf{0.40} & \textbf{0.10} & \cmark & {12.0s} [4.17it/s] \\
        \rowcolor{ourwhite2}
        Ours (w/o Delay) & {0.824} & 0.696 & \textbf{3.47} & 0.091 & 0.041 & \underline{0.50} & \underline{0.22} & \cmark & {12.0s} [4.17it/s] \\
        Ours but optimize $\|\tilde{\eps}_+ - \epspredtheta(\x_t\mid\emb+\Delta\emb)\|$ (no $\attributescalei$) &&& {2.23} &&& {{0.55}} & {{0.31}} & \cmark & {12.0s} [4.17it/s] \\
        \rowcolor{ourwhite2}
        Our CLIP Difference Method (\cref{sec:method_naive_deltas}) & - & - & 2.38 & - & - & 1.20 & 0.58 & \cmark & {12.0s} [4.17it/s] \\
        CLIP Delta without Difference: $\pembdiff_{\attributei} = (\pembp)_{[\mathrm{EOS}]}$ &&& 1.98 &&& 1.16 & 0.58 & \cmark & {12.0s} [4.17it/s] \\
        \rowcolor{ourwhite2}
        CLIP Delta without Difference: $\pembdiff_{\attributei} = (\pembp)_{[\subjectj]}$ &&& 1.83 &&& 1.20 & 0.60 & \cmark & {12.0s} [4.17it/s] \\
        Directly modulating $\Delta\epspred$ (\cref{sec:method_robust_deltas}) with CFG & - & - & 3.15 & - & - & 0.73 & 0.39 & \cmark & 23.0s [2.17it/s]\\
        \bottomrule
    \end{tabular}
    }
    \caption{Extended version of \cref{tab:main_quantitative} with additional ablations/baseline versions of our method.}
    \label{tab:additional_ablations}
\end{table}

\begin{figure}[H]
    \centering
    \includegraphics[width=0.3\linewidth]{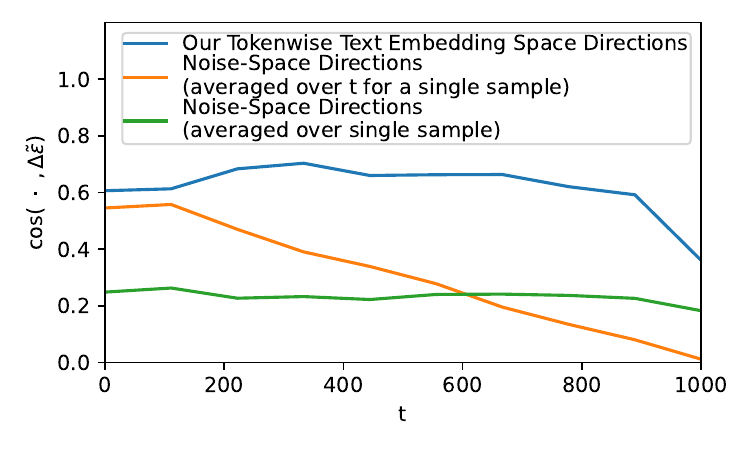}
    \caption{Cosine Similarities of approximations of $\Delta \tilde{\eps}$ compared to the actual true one over the diffusion timestep $t$.}
    \label{fig:cossims}
\end{figure}

\noindent We also compare with alternative approximations for the noise-space direction $\Delta \tilde{\eps}$ we are learning in the tokenwise text embedding space as $\pembdiff_{\attributei}$. Generally, other approaches to approximate these attribute- and sample-specific directions will not exhibit subject-specificity, so we perform this investigation in the single-subject case. We compare with two baselines that attempt to directly approximate $\Delta \tilde{\eps}$: averaging it over the diffusion timestep $t$ on a per-sample basis and averaging it over samples on a per-timestep basis. We compare them with the actual $\Delta \tilde{\eps}$ in \cref{fig:cossims}. We find that both of these approximations, despite still having a dependency on either $t$ or $\x_T$, only achieve a low similarity to the actual direction they attempt to approximate, while our directions $\pembdiff_{\attributei}$ consistently outperform both approximations over all $t$. 

\subsection{Challenging Attributes}
\begin{figure}[H]
    \centering
    \newcommand{\partspacing}{1.5mm}
    \newcommand{\imgheight}{.14\linewidth}
    \begin{minipage}{\linewidth}
        \adjustbox{max width=\linewidth}{
        \begin{tikzpicture}
            \node[simple node image] (reel1) {\includegraphics[height=\imgheight]{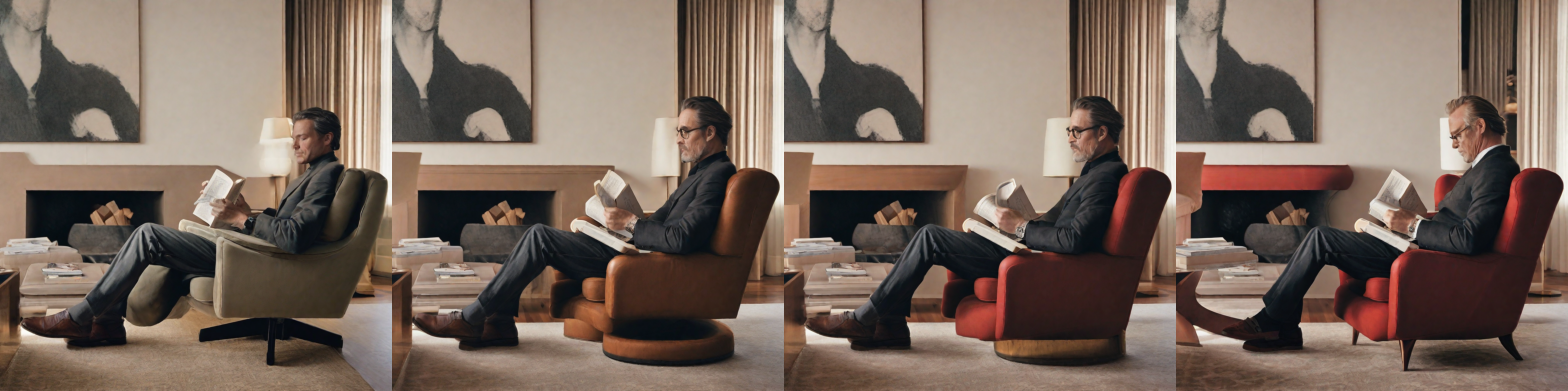}};
            \node[simple node image,right=\partspacing of reel1] (reel2) {\includegraphics[height=\imgheight]{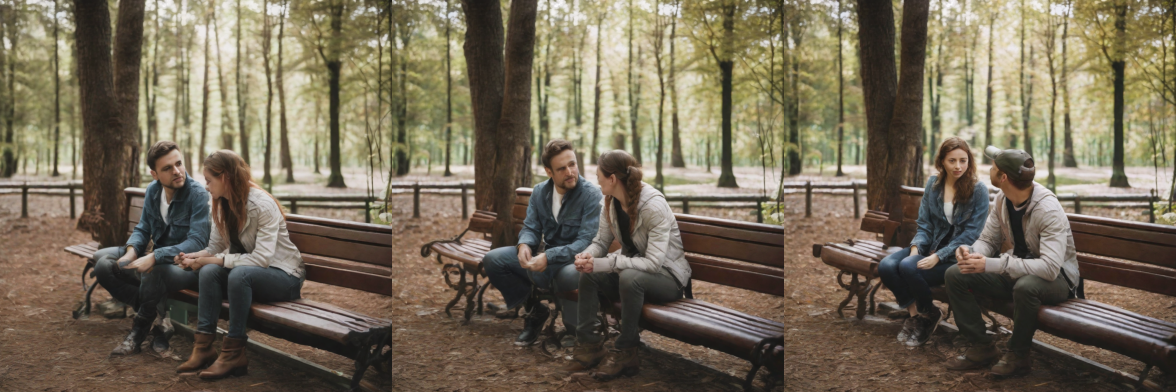}};
            \node[simple node image,right=\partspacing of reel2] (reel3) {\includegraphics[height=\imgheight,width=.42\linewidth]{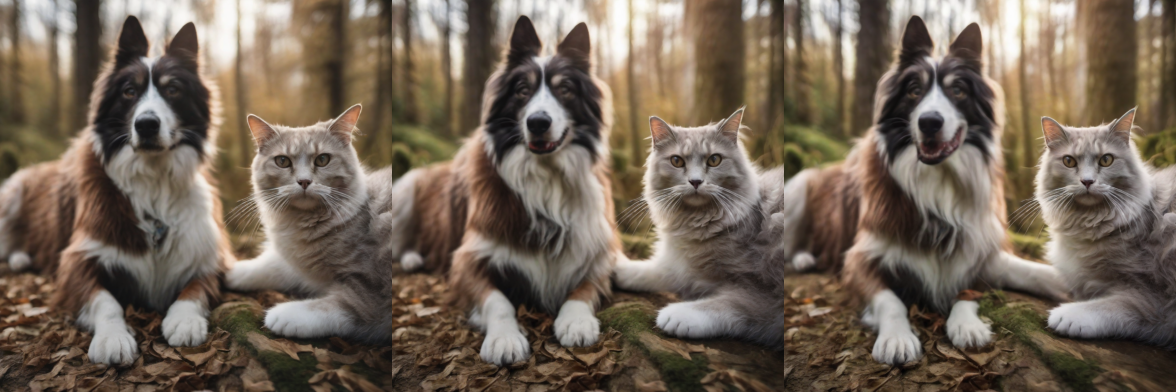}};
            
            \draw[standard arrow] ($(reel1.west |- reel1.north) + (0, 1mm)$) -- node[above,yshift=-1mm] {(a) chair: \textit{red} {(too far $\rightarrow$ leak)}} ($(reel1.east |- reel1.north) + (0, 1mm)$);
            \draw[standard arrow,draw=ourhighlightcolor] ($(reel1.east |- reel1.north) + (-7mm, -2mm)$) -- ($(reel1.east |- reel1.north) + (-8mm, -5mm)$);
            \draw[standard arrow] ($(reel2.west |- reel2.north) + (0, 1mm)$) -- node[above,yshift=-1mm] {(b) man: \textit{on the right}} ($(reel2.east |- reel2.north) + (0, 1mm)$);
            \draw[standard arrow] ($(reel3.west |- reel3.north) + (0, 1mm)$) -- node[above,yshift=-1mm] {(c) dog: \textit{happy}} ($(reel3.east |- reel3.north) + (0, 1mm)$);
            \node[draw=ourhighlightcolor,line width=0.04cm,below=0mm of reel1,minimum width=.143\linewidth,minimum height=.143\linewidth,yshift=.145\linewidth,xshift=-.21\linewidth] {};
            \node[draw=ourhighlightcolor,line width=0.04cm,below=0mm of reel2,minimum width=.143\linewidth,minimum height=.143\linewidth,yshift=.145\linewidth,xshift=-.14\linewidth] {};
            \node[draw=ourhighlightcolor,line width=0.04cm,below=0mm of reel3,minimum width=.143\linewidth,minimum height=.143\linewidth,yshift=.145\linewidth,xshift=-.14\linewidth] {};
            
        \end{tikzpicture}
        }
    \end{minipage}
    \caption{Vanilla Attribute Control in challenging settings.}
    \label{fig:challenging_settings}
\end{figure}

Some attributes are known in the community to be specifically challenging to get to work in practical settings. We show some successful examples of applying Attribute Control to them in \cref{fig:challenging_settings}. Color attributes (\cref{fig:challenging_settings}a) are known to be prone to leakage across different objects, even in the base model. Our method generally inherits these limitations from the base model and can not address cases where the original prompt already leads to attribute leakage. When \textit{adding} new attributes to the generated image, such as specifying the color for one object, we empirically find our modulations to lead to less (but still not zero) leakage. Intuitively, this makes sense, as we do not add an additional token describing the color change, which could be leaked to later tokens by the CLIP model and which any head of the diffusion model could attend to. Instead, we \textit{exclusively} add the information to the token that describes that object. However, as diffusion cross-attention maps are not fully leakage-free unless applying methods that deliberately enforce this~\cite{chefer2023attendexcite,rassin2023linguistic,xiao2024fastcomposer}, we still observe color leakage with attribute control, although to a lesser extent. This especially occurs when leakage is already present in the base generation or when too much control is exerted (as shown in \cref{fig:challenging_settings}a). Similarly, cases where the base model is prone to leakage (e.g., trying to affect dogs and cats separately) are less prone to attribute leakage when adding them via our method (see, e.g., \cref{fig:challenging_settings}c).
For attributes where the base model already struggles to apply them at all, our method inherits these limitations. Such attributes like spatial relations \textit{can} work (see \cref{fig:challenging_settings}b), but only do so (very) rarely, reflecting the base model's inability to parse them from normal prompts reliably.

\paragraph{Postfix Attribute Learning}\label{sec:app_postfix_attributes}
Some attributes are not easily expressible as prefixes to the noun. This means that, due to the causal nature of the CLIP text encoder, our optimization-free method for identifying attribute directions (see \cref{sec:method_naive_deltas}) can not be applied. However, we find that this limitation does not apply to our optimization-based approach (see \cref{sec:method_robust_deltas}): we can learn directions based on attributes expressed as postfixes (e.g., \textit{``a person \underline{wearing sunglasses}''}, for which we show a qualitative example in \cref{fig:sunglasses}).

\begin{figure}[H]
    \centering
    \adjustbox{max width=.4\linewidth}{\input{figs/sunglasses/figure}}
    \caption{Our learning-based method can also learn to represent attributes represented as postfixes to the target subject noun during training.}
    \label{fig:sunglasses}
\end{figure}

\subsection{Subject Noun Transferability}\label{sec:app_subject_noun_transferability}
We investigate how much our learned attribute modulations can generalize across different nouns that describe the same subject. We generally learn them on a set of different nouns that describe a subject of a specific category (e.g., for people with the words ``man'', ``woman'', and ``person''). However, these words typically do not cover the whole range of possible nouns that can be used to describe subjects of a general category. Ideally, one could learn one modulation for one concept, such as age, on a small set of nouns and generalize across all nouns of a category or even to subjects of other categories.

First, we test the generalization of modulations learned for people on ``man'', ``woman'', and ``person'' and apply them to increasingly more specific nouns that describe people. Results are shown in \cref{fig:noun_transfer_1,fig:noun_transfer_2}, and all prompts are ``a photo of a beautiful $<$noun$>$''. As a baseline, we apply them to ``child'', ``mother'', and ``father'', three words that are previously unseen but still describe very high-level sub-categories of people. We find that the learned modulations still work as expected. Similarly, for categories of jobs such as ``doctor'', ``barista'', or ``firefighter'', which are substantially more specific and also substantially affect their clothing and the rest of the image, we find that they also work well. Finally, applying these learned modulations to very specific nouns such as the names ``John'' and ``Jane'' also works as expected. This demonstrates that our learned modulations can generalize well across a wide range of unseen nouns describing instances of a specific category, even if they were only learned on a small set of high-level, potential nouns.

\begin{figure}[H]
    \centering
    {
        \scriptsize
        \adjustbox{max width=\linewidth}{\input{figs/appendix/transferability/age}}
    }
    \caption{\textbf{Subject Noun Transferability}. We stress-test applying modulations that have been learned only on the nouns ``man``, ``woman``, and ``person`` to various other nouns that describe people. The unmodified image is marked in \ourhighlightcolorstr. All samples are generated using attribute modulations being applied with a linear scale from -2 to 2 across each.}
    \label{fig:noun_transfer_1}
\end{figure}

\begin{figure}[H]
    \centering
    {
        \scriptsize
        \adjustbox{max width=\linewidth}{\input{figs/appendix/transferability/width}}
    }
    \caption{\textbf{Subject Noun Transferability}. We stress-test applying modulations that have been learned only on the nouns ``man``, ``woman``, and ``person`` to various other nouns that describe people. The unmodified image is marked in \ourhighlightcolorstr. All samples are generated using attribute modulations being applied with a linear scale from -2 to 2 across each.}
    \label{fig:noun_transfer_2}
\end{figure}

\subsection{Multi-Subject Attribute Editing}\label{sec:app_multi_subject_attribute_editing}
\cref{fig:app_multi_subject_1,fig:app_multi_subject_2} show examples of modulating attributes in a subject-specific manner using our learned modulations. These show that various attributes can be applied to subjects individually, even if both subjects are of the same category (e.g., ``people''). A slight correlation between, e.g., the age of the man and the age of the woman in \cref{fig:app_multi_subject_1} is visible and expected, as the diffusion model also models these dependencies between different subjects in the generated image. By applying both modulations with different strengths, the whole spectrum of combinations can be achieved, as shown in \cref{fig:age_coverage}.

\begin{figure}[H]
    \centering
    \adjustbox{max width=.7\linewidth}{\input{figs/appendix/multi_subject/age}}
    \adjustbox{max width=.7\linewidth}{\input{figs/appendix/multi_subject/width_2}}
    \caption{\textbf{Multi-Subject Attribute Modifications}. The unmodified image is marked in \ourhighlightcolorstr. All samples are generated using one attribute modulation each being applied to the two subjects mentioned in the prompt with a linear scale from -2 to 2 across each.}
    \label{fig:app_multi_subject_1}
\end{figure}

\begin{figure}[H]
    \centering
    \adjustbox{max width=.7\linewidth}{\input{figs/appendix/multi_subject/width}}
    \adjustbox{max width=.7\linewidth}{\input{figs/appendix/multi_subject/age_2}}
    \caption{\textbf{Multi-Subject Attribute Modifications}. The unmodified image is marked in \ourhighlightcolorstr. All samples are generated using one attribute modulation each being applied to the two subjects mentioned in the prompt with a linear scale from -2 to 2 across each.}
    \label{fig:app_multi_subject_2}
\end{figure}

\clearpage
\subsection{Compositional Attribute Editing}
We show some 2d grids where two attributes are modulated for the same target subject in an additive manner in \cref{fig:app_compositional_1,fig:app_compositional_2}. Both attribute modulations interact with each other according to the world knowledge of the diffusion model to produce a realistic image for every combination.

\begin{figure}[H]
    \centering
    \adjustbox{max width=.56\linewidth}{\input{figs/appendix/double_attribute/car_age_price_2}}
    \adjustbox{max width=.56\linewidth}{\input{figs/appendix/double_attribute/car_age_price}}
    \caption{\textbf{Compositional Attribute Modifications}. The unmodified image is marked in \ourhighlightcolorstr. All samples are generated using two attribute modulations being applied additively with a linear scale from -2 to 2 across each.}
    \label{fig:app_compositional_1}
\end{figure}

\begin{figure}[H]
    \centering
    \adjustbox{max width=.7\linewidth}{\input{figs/appendix/double_attribute/width_smile}}
    \adjustbox{max width=.7\linewidth}{\input{figs/appendix/double_attribute/width_smile_2}}
    \caption{\textbf{Compositional Attribute Modifications}. The unmodified image is marked in \ourhighlightcolorstr. All samples are generated using two attribute modulations being applied additively with a linear scale from -2 to 2 across each.}
    \label{fig:app_compositional_2}
\end{figure}

\clearpage
\subsection{Continuous Attribute Modulation}
To illustrate the breadth of attributes that can be modulated and how continuous the attribute changes are, we show a range of attributes being continuously modulated. \cref{fig:app_continuous_delay_0,fig:app_continuous_delay_1,fig:app_continuous_delay_2,fig:app_continuous_delay_3} show examples where attribute modulations are applied with our delayed sampling, \cref{fig:app_continuous_full_1} shows attribute modulations applied for the full sampling time. For every category, we re-use the same sample instances as a starting point.
\begin{figure}[H]
    \centering
    \adjustbox{max width=\linewidth}{\input{figs/appendix/single_attribute/vehicle_age}}
    \adjustbox{max width=\linewidth}{\input{figs/appendix/single_attribute/furniture_age}}
    \caption{\textbf{Continuous Attribute Modifications}. Unmodified images are marked in \ourhighlightcolorstr. All samples are generated using a linear scale from -2 to 2.}
    \label{fig:app_continuous_delay_0}
\end{figure}
\begin{figure}[H]
    \centering
    \adjustbox{max width=.85\linewidth}{\input{figs/appendix/single_attribute/age}}
    \adjustbox{max width=.85\linewidth}{\input{figs/appendix/single_attribute/fitness}}
    \adjustbox{max width=.85\linewidth}{\input{figs/appendix/single_attribute/tired}}
    \caption{\textbf{Continuous Attribute Modifications}. Unmodified images are marked in \ourhighlightcolorstr. All samples are generated using a linear scale from -2 to 2.}
    \label{fig:app_continuous_delay_1}
\end{figure}
\begin{figure}[H]
    \centering
    \adjustbox{max width=.85\linewidth}{\input{figs/appendix/single_attribute/elegant}}
    \adjustbox{max width=.85\linewidth}{\input{figs/appendix/single_attribute/freckled}}
    \adjustbox{max width=.85\linewidth}{\input{figs/appendix/single_attribute/groomed}}
    \caption{\textbf{Continuous Attribute Modifications}. Unmodified images are marked in \ourhighlightcolorstr. All samples are generated using a linear scale from -2 to 2.}
    \label{fig:app_continuous_delay_2}
\end{figure}
\begin{figure}[H]
    \centering
    \adjustbox{max width=.85\linewidth}{\input{figs/appendix/single_attribute/makeup}}
    \adjustbox{max width=.85\linewidth}{\input{figs/appendix/single_attribute/pale}}
    \adjustbox{max width=.85\linewidth}{\input{figs/appendix/single_attribute/width}}
    \caption{\textbf{Continuous Attribute Modifications}. Unmodified images are marked in \ourhighlightcolorstr. All samples are generated using a linear scale from -2 to 2.}
    \label{fig:app_continuous_delay_3}
\end{figure}

\begin{figure}[H]
    \centering
    \adjustbox{max width=.85\linewidth}{\input{figs/appendix/single_attribute_no_delay/long_hair}}
    \adjustbox{max width=.85\linewidth}{\input{figs/appendix/single_attribute_no_delay/scarred}}
    \adjustbox{max width=.85\linewidth}{\input{figs/appendix/single_attribute_no_delay/curly_hair}}
    \caption{\textbf{Continuous Attribute Modifications}. Unmodified images are marked in \ourhighlightcolorstr. All samples are generated using a linear scale from -2 to 2, with the modulations being applied \underline{for all steps} (w/o Delay).}
    \label{fig:app_continuous_full_1}
\end{figure}

\section{Implementation Details}
This section gives details about the implementation of our method.
We generally use the default settings as set in \texttt{diffusers}\footnote{\url{https://github.com/huggingface/diffusers}}-v0.25.0 with a classifier-free guidance \citep{ho2021classifierfree} scale of 7.5 and 50-step DDIM \citep{song2021ddim} sampling unless specified otherwise.

\subsection{Semantic Direction Training}\label{sec:app_training_details}

\begin{algorithm*}[ht]
\caption{Algorithm for Learning the Semantic Directions}\label{alg:semantic_directions}
\begin{algorithmic}[1]
    \State \textbf{Input:} 
        \Statex \hspace{1em} Pre-trained diffusion model $\epspredtheta$
        \Statex \hspace{1em} CLIP embedding dimension $d_\mathrm{CLIP}$
        \Statex \hspace{1em} Learning rate $\eta$, number of steps $S$, batch size $B$
    \State \textbf{Output:} 
        \Statex \hspace{1em} Learned semantic direction $\pembdiff_{\attributei}$
    
    \State Initialize $\pembdiff_{\attributei} = \mathbf{0} $ \Comment{Initialization}
    
    \For{$s = 1$ to $S$} \Comment{Training loop}
        \State $\mathcal{L}_{\text{batch}} \gets 0$ \Comment{Initialize batch loss}
        \For{each entry in batch of size $B$}
            \State Sample random subject $\subjectj$ and neutral prompt $\prompt$
            \State Generate image $\xzero$ from neutral prompt $\prompt$
            \State $t \sim \mathcal{U}[0, T]$ \Comment{Sample random timestep}
            \State $\x_t = \alpha_t \xzero + \sigma_t \eps, \eps \sim \mathcal{N}(0, \mathbf{I})$ \Comment{Add noise}
            
            \State $\epspred = \epspredtheta(\x_t|\prompt)$ \Comment{Predict noise for $P$}
            \State $\epspred_+ = \epspredtheta(\x_t|\promptp)$ \Comment{Predict noise for $P_+$}
            
            \State $\Delta \epspred = \epspred_+ - \epspred$ \Comment{Compute noise direction}

            \State $\attributescalei \sim \mathcal{U}([-5, 5] \setminus (-0.1, 0.1))$ \Comment{Sample scale factor}
            
            \State $\loss_i = w(t) \left\|(\eps + \attributescalei \Delta \epspred) - \epspredtheta(\x_t|\pembmod(\emb,\attributescalei\pembdiff_{\attributei}),t)\right\|_2^2$  \Comment{Compute loss for this entry}
            
            \State $\mathcal{L}_{\text{batch}} \gets \mathcal{L}_{\text{batch}} + \loss_i$ \Comment{Accumulate batch loss}
        \EndFor
        \State Compute mean loss for the batch: $\mathcal{L}_{\text{mean}} \gets \frac{1}{B} \mathcal{L}_{\text{batch}}$
        \State Update $\pembdiff_{\attributei}$ using AdamW optimizer with learning rate $\eta$ based on $\mathcal{L}_{\text{mean}}$
    \EndFor
    
    \State \textbf{Return:} $\pembdiff_{\attributei}$
\end{algorithmic}
\end{algorithm*}

The semantic directions $\pembdiff_{\attributei}$ for target attribute $\attributei$ are implemented as learnable parameters of shape $1 \times d_\mathrm{CLIP}$, with $d_\mathrm{CLIP}$ being the embedding dimension of the CLIP text encoder. For SDXL \citep{podell2024sdxl}, this is 2048, resulting from the channelwise concatenation of embeddings from the OpenAI CLIP ViT-L \citep{clip} and OpenCLIP ViT-bigG \citep{openclip_software}. This direction is applied additively with scaling according to \cref{eq:attribute_modulation} to the target subject tokens (e.g., ``person'' in the case of ``a photo of a person'') in the original text embedding $\emb$. If the target subject consists of multiple tokens, we broadcast $\pembdiff_{\attributei}$ across those tokens, although this is only very rarely the case in practice. Similarly, if one subject is mentioned in the prompt multiple times, we apply the same modulation to all instances.

We train our semantic directions $\pembdiff_{\attributei}$ for 1000 steps\footnote{The directions tend to be mostly converged after 10 steps, but we train for a unified training time across all attributes for consistency.} at a batch size of 10. We use AdamW \citep{loshchilov2018decoupled} with a learning rate of 0.1, $(\beta_1, \beta_2) = (0.5, 0.8)$, and weight decay of 0.333. All directions are trained on a single A100 with 40GB of VRAM using a bfloat16 version of SDXL \citep{podell2024sdxl}.

For every entry in the batch, we use a random combination of prefix prompt (e.g. ``an photo of", optionally with attributes such as ethnicity (e.g., \{asian, african-american, caucasian, arab, african, south-american, indian, ...\}), to focus the implied direction on one that is invariant to these attributes) and prompt tuple (e.g ``a woman") and sample an image with the neutral prompt (e.g. (``a photo of a woman") and a random seed, stopping at a random timestep. We then compute the prediction starting from that step for all two/three prompts, resulting in $\epspred, \epspred_+$, and optionally $\epspred_-$. In contrast to \cite{gandikota2023sliders}, we explicitly distill the full direction implied by $\Delta \epspred$ by using multiple scales $\attributescalei$ sampled from a continuous scale distribution. Preliminary experiments showed that this helps obtain substantially more robust directions. Additionally, we sample our starting samples using standard sampling instead of a modified generation process.

We then sample four values for $\attributescalei \sim \mathcal{U}([-5, 5]\setminus (-0.1,0.1))$ and compute our training loss (\cref{eq:loss_direction}) over them. We found that sampling multiple values for $\attributescalei$ substantially boosts the quality of our learned directions at little overhead cost (as the online sampling of the original images is the most costly part) and that values for $\attributescalei$ very close to zero were not particularly useful for the training process. Empirically, we find that most of our learned directions are already close to convergence after five optimization steps, but we keep training for the full time for simplicity.

\subsection{Combination of Attribute Control with other Methods}\label{sec:app_combination_attribute_control_other_methods}
In \cref{sec:experiments}, we combine our attribute control method with other off-the-shelf controlled generation methods.

\paragraph{Combination with Prompt-to-Prompt~\citep{hertz2023prompttoprompt}} To combine our method with Prompt-to-Prompt, we apply the standard Prompt-to-Prompt method. We use the same adaptation mode and hyperparameters as used for adding adjectives in the text prompt, but add our modulations on the text prompt embedding instead. To modulate the change, we scale our directions as usual.

\paragraph{Combination with AdapEdit~\citep{ma2024adapedit}} AdapEdit uses the same general external interface as Prompt-to-Prompt. Here, we apply our modulations in the exact same way as previously described for Prompt-to-Prompt. As AdapEdit is not available for SDXL~\citep{podell2024sdxl}, we use zero-shot adaptation of our semantic directions obtained on SDXL to SD1.5, as described in \cref{sec:exp_generalization}.

\paragraph{Combination with ReNoise~\citep{garibi2024renoise}} To apply our controlled generation approach to editing, we combine it with ReNoise, a standard inversion approach. We use their official reference implementation based on SDXL Turbo~\citep{sauer2023adversarial} and apply our modulations learned on SDXL there. We perform inversion purely with ReNoise with default settings and an image description prompt to obtain a starting latent $\x_T$, and then perform controlled generation purely with our method with standard settings. This could optionally be combined further with other methods during inference, such as Prompt-to-Prompt~\citep{hertz2023prompttoprompt} and AdapEdit~\citep{ma2024adapedit}.

\subsection{Experiment Evaluation Details}\label{sec:app_experiment_evaluation_details}
To compute perceptual image differences, we use LPIPS \citep{lpips} as implemented in the \texttt{lpips}\footnote{\url{https://github.com/richzhang/PerceptualSimilarity}} package with default settings at a resolution of $256^2$ (interpolated bi-linearly). For CLIP scores, we use the standard implementation in \texttt{torchmetrics}\footnote{\url{https://github.com/Lightning-AI/torchmetrics}} (which outputs cosine similarities scaled to $[0, 100]$) with default settings, including the default CLIP choice of the CLIP-ViT-L/14 trained by OpenAI \citep{clip}. For image-image similarity evaluations with DINOv2 \citep{dinov2}, we use the ViT-L/14 variant with registers \citep{registers} and bi-linearly resize to $224^2$ before passing them to the model and comparing the cosine similarity of the CLS token outputs. Finally, for ReID evaluations, we use the ArcFace \citep{deng2019arcface} implementation provided by the \texttt{insightface}\footnote{\url{https://github.com/deepinsight/insightface}} python package with the default \texttt{buffalo\_l} model, where we compute the cosine similarity of the embeddings of the detected faces.

\paragraph{Implementations of other Methods}
For Concept Sliders~\citep{gandikota2023sliders}, we use the official public implementation\footnote{\url{https://github.com/rohitgandikota/sliders}}. For Prompt-to-Prompt~\citep{hertz2023prompttoprompt}, we use RoyiRa's unofficial port of the method to Stable Diffusion XL\footnote{\url{https://github.com/RoyiRa/prompt-to-prompt-with-sdxl}}. This implementation also served as the basis for integrating our method with Prompt-to-Prompt in our codebase. As this implementation is partially incomplete, we referred to the official implementation Prompt-to-Prompt\footnote{\url{https://github.com/google/prompt-to-prompt}} for the implementation of reweighting of added words. For AdapEdit\footnote{\url{https://github.com/AnonymousPony/adap-edit}}, MasaCtrl\footnote{\url{https://github.com/TencentARC/MasaCtrl}}, and ReNoise\footnote{\url{https://github.com/garibida/ReNoise-Inversion}}, we also used the respective official implementations.
When comparing attribute modulation capabilities across different methods, we compare using the target attribute age on people, as this attribute is i) unambiguous in what exactly it describes, ii) fully continuous, and iii) the attribute supported by Concept Sliders\footnote{\url{https://sliders.baulab.info/weights/xl_sliders/}} that can be evaluated most objectively while being one that SD(XL) can readily interpret when given as text (unlike, e.g., eye size).

\paragraph{Attribute Distribution Shifts (\Cref{fig:attribute_distribution_shift_kde})}
For each value of $\attributescalei \in \{0, 1, 2, 3\}$, 20 samples (with fixed seeds across scales) were drawn. We compute the delta CLIP score as specified in the experiments section of the paper and use scipy's Gaussian KDE method\footnote{\url{https://docs.scipy.org/doc/scipy/reference/generated/scipy.stats.gaussian_kde.html}} to compute the kernel density estimate for the resulting distributions with Scott's rule and default settings.

\paragraph{Qualitative Continuous Modulation (\Cref{fig:age_qualitative})}
We continuously modulate the age of the person described in the prompt with both our method and Concept Sliders~\citep{gandikota2023sliders}, choosing coefficients such that a wide range is covered and both methods show similar scales per column. For Prompt-to-Prompt~\citep{hertz2023prompttoprompt} and MasaCtrl~\citep{cao2023masactrl}, we add ``old'' or ``young'' to the prompt to coarsely modulate the target attribute. Prompt-to-Prompt further enables some fine-grained control \textit{around the already offset attribute expression point from the added adjective} by re-weighting the added adjective. This does, at least for Stable Diffusion XL~\citep{podell2024sdxl}, not allow continuous modulation back to the original image, causing a discontinuity. This can intuitively be explained by the fact that attributes are aggregated in the subject noun, a fact that our method exploits to directly enable fine-grained, subject-specific target attribute modulation: as the attribute modulation for P2P is already partially contained in the subject noun, modulating just the added adjective's cross-attention map can not fully recover the original generated image. At the same time, when combined with our method, where we just modulate the target subject noun's embedding instead of adding new adjectives, this problem immediately subsides.

\paragraph{Quantitative Subject Specificity Evaluation (\Cref{tab:main_quantitative}a)}
With each method, we generate variations across a set of 50 images with individual prompts describing two people, where we modulate the target attribute of one of the two subjects. We detect each subject in the unmodified image as previously described with the standard pipeline from \texttt{insightface}, and then compute the target metric for each bounding box. We aggregate the specificity metric as described in \cref{eq:specificity} by computing the fraction individually per sample and then aggregating the overall mean. As there are some cases where this effectively results in a division by zero, we clamp the resulting individual values to $[0, 10]$. We chose 10 as a threshold, as it prevents these outlier samples from having an extraordinarily strong effect on the overall mean.

\paragraph{Attribute Coverage Evaluation (\Cref{fig:age_coverage})}
To evaluate the set of attribute combinations reachable by each method, we start from the same setup as previously described for \Cref{tab:main_quantitative}a, but continuously modulate the age for both subjects visible in the image, covering all combinations of modulation scales for each method. We evaluate 20 values per subject, producing 400 generated samples per method for methods that allow independent continuous modulation of both subjects. We then measure the attribute expression for each subject bounding box (obtained as previously in \Cref{tab:main_quantitative}a) using \cref{eq:delta_clip_bi} and plot the distribution for one representative sample in \cref{fig:age_coverage}.

\paragraph{Quantitative Disentangledness Evaluation (\Cref{fig:continuous_modulation_image_change}, \Cref{tab:main_quantitative}b)}
We generate 50 base samples showing people with different prompts of the format \textit{``a close-up portrait of a \{modifiers\} \{woman, man\}''}, where \{modifiers\} describes a set of prefixes (e.g., \textit{``\{$\emptyset$, beautiful, elegant\} asian''}, \textit{``\{$\emptyset$, beautiful, elegant\} african-american''}, etc) to cover a wide variety of different images. Then, we modulate the target attribute continuously using each method. We then measure the attribute expression change with \cref{eq:delta_clip_bi}, the image change with LPIPS, and the identity change as in \cref{eq:delta_id}. We aggregate these values over all 50 images per combination of method \& hyperparameters and then plot them in \cref{fig:continuous_modulation_image_change}. For $\Cref{tab:main_quantitative}b$, we compute the slope of these graphs (using the absolute value of $\Delta \mathrm{CLIP}_\mathrm{Bi}$ for the denominator, to account for the fact that the changes increase for positive values and one for negative values of $\Delta \mathrm{CLIP}_\mathrm{Bi}$) to quantify the disentangledness of the edits both from overall visual changes (LPIPS) and person identity changes ($\Delta \text{Id}$).

\paragraph{Inference Performance Evaluation (\Cref{tab:main_quantitative}d)}
For each method, we use the released implementations of each respective method with default settings and replicate the original environments as closely as possible, given the information documented by the authors. We measure inference times on the same Nvidia A100 SXM with 80GB of VRAM and document both the total time and (average) step time, as some methods use different step counts for sampling. For the main paper, we consolidate inversion and generation time if applicable. We exclude the time spent obtaining attribute deltas, as it is done once ahead of time and causes no overhead during inference/amortizes quickly when needing to train deltas for new attributes, similar to Concept Sliders~\cite{gandikota2023sliders}, where we also exclude slider training time due to the same reason.

\clearpage
\section{Visualization Details \& Prompts}\label{sec:app_visualization_details_prompts}
Generally, all examples in the paper use Stable Diffusion XL as introduced by \citet{podell2024sdxl} unless noted otherwise. In the following, we provide the prompts and, in the case of editing examples, image sources incl.\ licenses, used to generate the various qualitative examples presented in the paper.

\paragraph{\Cref{fig:teaser}}\label{sec:prompts:delta_vs_text_change}
Prompt: \textit{``A close-up photo of a man and a woman sitting on a bench.''}

\paragraph{\Cref{fig:continuousness_interpolation}}
Prompts: \textit{``a portrait of a beautiful car''}, \textit{``a portrait of a beautiful frog''}, and \textit{``a portrait of a beautiful suv''}.

\paragraph{\Cref{fig:random_subject_specific_deviation}}
Prompt: \textit{``a portrait of a beautiful woman with her beautiful dog''}.

\paragraph{\Cref{fig:naive_vs_learned}}
Prompt: \textit{``a photo of a car''}.

\paragraph{\Cref{fig:attribute_distribution_shift_kde}}
Prompt: \textit{``a photo of a car''}.%

\paragraph{\Cref{fig:age_qualitative}}
Base prompt: \textit{``a close-up portrait of a indian woman''}.

\paragraph{\Cref{fig:real_image_editing}}
Image 1 is a photo with the title \textit{``a red rolls royce parked in front of a building''} by Rico Reynaldi, obtained from Unsplash\footnote{\url{https://unsplash.com/photos/a-red-rolls-royce-parked-in-front-of-a-building-sAN11DGnjqk}}. The image is licensed under the Unsplash license\footnote{\url{https://unsplash.com/license}} and has been center-cropped for inversion.\\
Inversion Prompt: \textit{``a photo of a beautiful red car on the top deck of a parking garage with large buildings in the background, hazy weather with sunshine''}.\\
Image 2 is a photo by The Royal Society, obtained from Wikimedia\footnote{\url{https://commons.wikimedia.org/wiki/File:Demis_Hassabis_Royal_Society.jpg}}. The image is licensed under the Creative Commons Attribution-Share Alike 3.0 Unported license\footnote{\url{https://creativecommons.org/licenses/by-sa/3.0/deed.en}} and has been cropped to primarily show the person's head.\\
Inversion Prompt: \textit{``a photo of a man wearing glasses and a suit''}.

\paragraph{\Cref{fig:attribute_delta_composability}a}
Prompt: \textit{``a photo of a beautiful asian man''}.

\paragraph{\Cref{fig:attribute_delta_composability}b}
Prompt: \textit{``a portrait of a bearded man and a beautiful brunette woman''}.

\paragraph{\Cref{fig:various_deltas}}
Prompt 1: \textit{``a portrait of a beautiful chair''}.\\
Prompt 2: \textit{`` photo of an old car''}.\\
Prompt 3: \textit{``a portrait of a beautiful truck''}.\\
Prompt 4: \textit{``a photo of a beautiful man''}.

\paragraph{\Cref{fig:zero_shot_transfer}}
aMUSEd: \textit{``a photo of a beautiful man''}.\\
SD 1.5: \textit{``a headshot of a relaxed woman and a friendly man''}.

\paragraph{\Cref{fig:pixart_alpha}a}
Prompt: \textit{``a photo of a beautiful man''}
\paragraph{\Cref{fig:pixart_alpha}b}
Prompt: \textit{``a photo of a beautiful woman''}
\paragraph{\Cref{fig:pixart_alpha}c}
Prompt: \textit{``a close-up photo of a real beautiful man with his beautiful cat sitting in the forest, high detail, wide angle lens.''}

\paragraph{\Cref{fig:challenging_settings}a}
Prompt: \textit{``A close-up photo of a man sitting in a chair. He is leaning back and reading a book. A sofa is seen in the background. modern aesthetic, architectural digest.''}
\paragraph{\Cref{fig:challenging_settings}b}
Prompt: \textit{``A close-up photo of a man and a woman sitting on a bench. The setting is in the forest, high detail, wide angle lens''}
\paragraph{\Cref{fig:challenging_settings}c}
Prompt: \textit{``A close-up photo of a dog sitting next to a cat. The setting is in the forest, high detail, wide angle lens''}

\paragraph{\Cref{fig:sunglasses}} Prompt: \textit{``A photo of a beautiful asian man''}

\paragraph{\Cref{fig:noun_transfer_1,fig:noun_transfer_2}}
Prompt Template: \textit{``a photo of a beautiful [...]''}

\paragraph{\Cref{fig:app_multi_subject_1}}
Prompt 1: \textit{``a photo of a bearded man in a beanie enjoying a concert with a bohemian woman in flowing attire''}\\
Prompt 2: \textit{``a portrait of an indian woman standing next to an african-american man''}

\paragraph{\Cref{fig:app_multi_subject_2}}
Prompt 1: \textit{``a photo of a tech-savvy man with a laptop engaged in conversation with a creative woman with colorful tattoos''}\\
Prompt 2: \textit{``a portrait of an indian woman dressed in traditional clothing next to an african-american man wearing a hat standing in a library''}

\paragraph{\Cref{fig:app_compositional_1}}
Prompt 1: \textit{``a photo of a car''}\\
Prompt 2: \textit{``a photo of a compact red car''}

\paragraph{\Cref{fig:app_compositional_2}}
Prompt 1 \& 2: \textit{``a photo of a beautiful asian man''}

\paragraph{\Cref{fig:app_continuous_delay_0}}
Prompt 1 \& 2: \textit{``a photo of a bike''}\\ 
Prompt 3 \& 4: \textit{``a photo of a car''}\\
Prompt 5 \& 6: \textit{``a photo of a bed''}\\
Prompt 7 \& 8: \textit{``a photo of a chair''}

\paragraph{\Cref{fig:app_continuous_delay_1,fig:app_continuous_delay_2,fig:app_continuous_delay_3,fig:app_continuous_full_1}}
Prompt 1 \& 3: \textit{``a photo of a beautiful man''}\\
Prompt 2 \& 4: \textit{``a photo of a beautiful woman''}

\end{document}